\documentclass[conference]{IEEEtran}

\usepackage{booktabs} % For formal tables
\usepackage{verbatim}
\usepackage{subcaption}
\usepackage{pifont}% http://ctan.org/pkg/pifont
\usepackage[linesnumbered,ruled,vlined]{algorithm2e}
\SetKwInput{KwInput}{Input}
\usepackage{adjustbox}
\usepackage{amsmath}
\usepackage{hyperref}
\usepackage{tabularx}
\usepackage{multirow}
\SetKwInput{KwInput}{Input}

\usepackage{csquotes}
\usepackage[noabbrev,capitalise]{cleveref}
\newcount\colveccount
\newcommand*\colvec[1]{
        \global\colveccount#1
        \begin{pmatrix}
        \colvecnext
}
\def\colvecnext#1{
        #1
        \global\advance\colveccount-1
        \ifnum\colveccount>0
                \\
                \expandafter\colvecnext
        \else
                \end{pmatrix}
        \fi
}

\def\BibTeX{{\rm B\kern-.05em{\sc i\kern-.025em b}\kern-.08em
    T\kern-.1667em\lower.7ex\hbox{E}\kern-.125emX}}
\begin{document}

\title{A Case Study on Optimization of Warehouses}

\author{\IEEEauthorblockN{Veronika Lesch}
\IEEEauthorblockA{\textit{University of Würzburg}\\
Würzburg, Germany \\
veronika.lesch@uni-wuerzburg.de}
\and
\IEEEauthorblockN{Patrick B.M. Müller}
\IEEEauthorblockA{\textit{University of Applied Sciences Würzburg}\\
Würzburg, Germany\\
patrick.mueller@fhws.de}
\and
\IEEEauthorblockN{Moritz Krämer}
\IEEEauthorblockA{\textit{io-consultants GmbH \& Co. KG}\\
Heidelberg, Germany}
\and
\IEEEauthorblockN{Samuel Kounev}
\IEEEauthorblockA{\textit{University of Würzburg}\\
Würzburg, Germany\\
samuel.kounev@uni-wuerzburg.de}
\and
\IEEEauthorblockN{Christian Krupitzer}
\IEEEauthorblockA{\textit{University of Hohenheim}\\
Hohenheim, Germany \\
christian.krupitzer@uni-hohenheim.de}
}

\maketitle

\begin{abstract}
In warehouses, order picking is known to be the most labor-intensive and costly task in which the employees account for a large part of the warehouse performance.
Hence, many approaches exist, that optimize the order picking process based on diverse economic criteria. 
However, most of these approaches focus on a single economic objective at once and disregard ergonomic criteria in their optimization.
Further, the influence of the placement of the items to be picked is underestimated and accordingly, too little attention is paid to the interdependence of these two problems.  
In this work, we aim at optimizing the storage assignment and the order picking problem within mezzanine warehouse with regards to their reciprocal influence.
We propose a customized version of the Non-dominated Sorting Genetic Algorithm II (NSGA-II) for optimizing the storage assignment problem as well as an Ant Colony Optimization (ACO) algorithm for optimizing the order picking problem. 
Both algorithms incorporate multiple economic and ergonomic constraints simultaneously.
Furthermore, the algorithms incorporate knowledge about the interdependence between both problems, aiming to improve the overall warehouse performance.
Our evaluation results show that our proposed algorithms return better storage assignments and order pick routes compared to commonly used techniques for the following quality indicators for comparing Pareto fronts: Coverage, Generational Distance, Euclidian Distance, Pareto Front Size, and Inverted Generational Distance.
Additionally, the evaluation regarding the interaction of both algorithms shows a better performance when combining both proposed algorithms.
\end{abstract}


\section{Introduction}
Warehouses play a central role in the supply chain of a company and contribute to its logistical success.
When employing humans, picker-to-parts and parts-to-picker methods are differentiated~\cite{Koster2007}.
Experts estimate the picker-to-parts system to be the most common in Western Europe with a share of over~80\%~\cite{Dekoster2007}.
A well-known picker-to-parts system is the mezzanine warehouse which we address in this work.

Working within a mezzanine warehouse consists of two main tasks: (i)~filling the storage with goods~(storage assignment) and (ii)~picking items out of the storage~(order picking).
The storage assignment problem defines the task of selecting storage locations to put a product into storage.
The order picking problem defines the task of computing a pick route that collects the requested products of a customer order.
Finding suitable storage allocations is important, as the allocation of products affects the travel distances during order picking.
Due to the NP-hardness and, hence, the complexity of the storage assignment and the order picking problem, efficient optimization algorithms are required to find satisfying solutions within acceptable times.
In the literature, many approaches exist for optimizing both warehouse problems.
However, most approaches usually target either of the warehouse problems; some works target both problems, however miss to integrate the interrelation between them and view each problem separately~\cite{VanGils2018}.
However, as identified by~\cite{Gu2010}, warehouse problems are strongly coupled. 
Thus, optimizing each warehouse problem individually may yield suboptimal solutions, harming the overall warehouse performance.
Since the employees spend most time traveling in such a mezzanine warehouse~\cite{Dekoster2007}, it is not surprising that most approaches focus on optimizing the travel distance. 
Additionally, ergonomic constraints are rarely considered, even though mezzanine warehouses represent labor-intensive working environments.

In this paper, we propose an integrated approach for combined storage assignment and order picking that simultaneously optimizes multiple economic and ergonomic constraints in mezzanine warehouses.
Expert interviews have shown, that in practice the following set of economic criteria is important and, hence, supported by our approach: products should be spread equally among each floor, fast-moving products should be easily accessible, correlated products should be stored in proximity of each other, and the storage space should be used as efficiently as possible.
Further, we integrate ergonomic constraints such as storing heavy products and fast-moving products at grip height or reducing the requirement to switch a mezzanine floor.
In an evaluation using three simulated mezzanine warehouses of different sizes, we analyze the quality of the solutions returned by our algorithms compared to commonly used techniques.
Finally, we assess the quality improvement when combining both of our algorithms compared to an isolated application.
Hence, the contribution of this paper is threefold:
\begin{enumerate}
    \item Design of storage allocation and order picking algorithms that incorporate the interdependence of both tasks. 
    \item Integration of diverse economic and ergonomic constraints.
    \item Evaluation of the approach in a use case based on real-world data provided by our cooperation company.
\end{enumerate}

The remainder of this paper is structured as follows.
Section~\ref{sec:relwork} presents related work and delineates our paper from existing approaches.
Section~\ref{sec:metamodel} presents the meta-model and floor layout of considered mezzanine warehouses.
Afterwards, Section~\ref{sec:storageassignment} provides an overview of the goal and a 3-phase algorithm of our storage assignment approach, while Section~\ref{sec:sa_phase2} presents the details of the proposed Genetic Algorithm for storage assignment.
Then, Section~\ref{sec:orderpicking} shows our order picking approach based on an adapted Ant Colony Optimization algorithm.
Section~\ref{sec:eval} presents our evaluation methodology and discusses the results and threats to validity.
Finally, Section~\ref{sec:conclusion} concludes the paper and summarizes future work.

\section{Related Work}
\label{sec:relwork}
In the literature, diverse storage assignment policies exist such as the dedicated and the random storage policy~\cite{Bartholdi2019}, the closest open location storage policy~\cite{Dekoster2007}, rank-based storage policies~\cite{Petersen1999}.
Further, class-based, golden zone, and family grouping storage policies are introduced in the literature~\cite{Dekoster2007,Petersen2005}.
Additionally, diverse approaches apply optimization techniques.
\cite{Sooksaksun2012} propose a particle swarm optimization algorithm for warehouses that deploy the class-based storage policy.
\cite{Kovacs2011} presents a mixed integer programming model for optimizing the storage assignment problem for class-based assigned warehouses.
\cite{Kofler2010} apply local search algorithms for reorganizing the products in the warehouse to keep it operating efficiently. 
\cite{Li2008} propose a multi-objective genetic algorithm for optimizing the storage assignment problem in automated storage/retrieval warehouses.

Similarly, heuristic policies exist for the order picking problem such as the S-Shape, Return, Mid-Point, Largest Gap, and Combined heuristic \cite{Petersen1997,Roodbergen2001,Vaughan1999}
Besides, \cite{Ratliff1983} presents an optimal algorithm using dynamic programming to find the shortest pick route in a single-block warehouse.
Additionally, \cite{Daniels1998} propose a mathematical model in combination with construction heuristics and apply Tabu Search to construct order picking routes.
\cite{Ene2012} present an integer programming model for optimizing the order picking problem.
\cite{Xing2010} propose an Max-Min Ant System~(MMAS) algorithm for optimizing machine travel paths in automated storage/retrieval warehouses. 
\cite{Chen2013} propose an ACO algorithm that detects congestion situations that arise when multiple order pickers traverse the same pick aisle simultaneously.

Finally, related work also assess the interaction of storage assignment and order picking approaches.
\cite{Petersen1999} and \cite{VanGils2018} provide an overview of well-performing combinations of storage assignment strategies and routing heuristics.
\cite{Manzini2007} analyze different parameters that affect the travel time in single-block warehouses that deploy the class-based storage policy.
\cite{Shqair2014} study the effects of different parameters on the travel distance in multi-block warehouses.

Our work delineates from these existing approaches in diverse aspects.
First of all, our work applies optimization techniques and does not rely on a policy on how to select fitting storage racks or shortest pick routes.
Second, regarding existing optimization approaches, our work integrates multiple objectives at once considering economic as well as ergonomic constraints at once while most of the other approaches focus on a single economic goal.
Finally, in contrast to existing work that address the influence of storage assignment and order picking tasks, we designed algorithms that optimize the targets of both tasks.
Hence, they optimize storage assignment and order picking with regards to the interdependence of both algorithms, while other works only provide well-performing combinations of algorithms or perform parameter tuning.

\section{Foundations on Optimization}
\label{sec:found:optprob}
Mathematically, an optimization problem can be defined as~\cite{rao2019engineering}:
\begin{equation}
    \text{Find } X = \colvec{4}{x_1}{x_2}{\vdots}{x_n} \text{ which minimizes } f(X)
\end{equation}
subject to the constraints
\begin{align}
    g_j(X) &\leq 0, \qquad j = 1,2,\dots, m\\
    l_j(X) &= 0, \qquad j = 1,2,\dots, p
\end{align}
where $X$ is the design vector with n dimensions comprising the variables to be determined by the optimization process.
$f(X)$ is the objective function and $g_j(X)$ and $l_j(X)$ are inequality and equality constraints, respectively.
Constraints limit the range of values to which the design variables can be set, and, thus, represent functional and other requirements on the solution.
Constraints can be classified into two types: (i)~hard constraints and (ii)~soft constraints. 
While hard constraints must be satisfied to find a feasible solution, the soft constraints should be satisfied, and any failure to satisfy with this constraint is penalized in the objective function. 
A feasible solution is called locally optimal if the neighborhood of this solutions does not contain solutions with better objective function values.
A feasible solution is called globally optimal if all possible solutions in the design space achieve lower objective function values.
The goal of optimization processes and optimization algorithms is to find the globally optimal solution.
When the complexity of an optimization problem does not allow finding the globally optimal solution, an optimization algorithm returns the best solution found so far.
According to \cite{rao2019engineering}, optimization problems can be classified into several categories: based on the existence of constraints, based on the nature of design variables, based on the physical structure of the problem, based on the nature of equations involved, based on the permissible values of the design variables, based on the deterministic nature of variables, based on the separability of the functions, and based on the number of objective functions.

\subsection{Multi-objective Optimization}
\label{sec:found:moo}
When considering optimization problems, multiple and even conflicting objectives must often be considered~\cite{Ngatchou2005Pareto}.
Traditionally, these objectives are integrated into a single objective function by aggregating all objectives with predefined weights or converting them into constraints. 
According to \cite{Ngatchou2005Pareto}, this leads to four limitations.
First, defining the aggregated objective function requires a priori knowledge about the importance of each objective.
Second, aggregating the objectives into a single function leads to a single solution.
Third, this leads to the impossibility to balance the objectives in a set of solutions.
Fourth, it is possible that a solution cannot be obtained unless the search space is convex.
Therefore, this process of objective reduction is not feasible for complex multi-objective optimization problems, leading to the concept of Pareto fronts.
Pareto optimality theory aims to balance a set of possibly conflicting objectives to find a number of solutions that perform equally well~\cite{Wang2016}: 
\enquote{Pareto optimality defines dominance to compare solutions, i.e., a solution A is said to dominate another solution B, if for all objectives, A is no worse than B at the same time at least one objective exists that A is better than B.}
The resulting set of solutions that cannot be dominated by other solutions is called Pareto front. 
The set of all non-dominated solutions that exist in a design space are called optimal Pareto front, while the Pareto front determined by an optimization algorithm is called the computed Pareto Front~\cite{Wang2016,Ngatchou2005Pareto}.

\subsection{Quality Indicators for Multi-objective Optimization}
\label{sec:found:hv}
To assess the quality of a Pareto front, we summarize the following performance indicators as introduced by~\cite{Wang2016}.
These metrics use the concept of a reference Pareto front ($PF_{ref}$) or an optimal Pareto front.
The performance indicators can be categorized into four performance categories: (i)~the Coverage aspect evaluated using C, (ii)~Convergence evaluated using GD and ED, (iii)~Diversity evaluated using PFS and GS, and (iv)~Combination evaluated by IGD and HV.

First, the C performance indicator defined in \cref{eq:eval:wh:c} quantifies the extent to which a computed Pareto front~($PF_c$) covers the reference Pareto front, i.e., the number of solutions~($s$) of the computed Pareto front that are also part of the reference Pareto front divided by the number of solutions in the reference Pareto front.
The best value for this metric is one since this indicates that the computed Pareto front covers the whole reference Pareto front which represents the best known solutions.
\begin{equation}
    \label{eq:eval:wh:c}
    C= \frac{|\cup_{s\in PF_c} s \in PF_{ref}|}{|PF_{ref}|}
\end{equation}

Second, the GD performance indicator measures the Euclidean distance from each solution in the computed Pareto front to the nearest solution in the reference Pareto front~($d(s_i, PF_{ref})$) and is defined in \cref{eq:eval:wh:gd}.
Hence, it provides a measure to judge how much distance is present between the computed and the best known solution.
The best possible result for this metric is zero, as it indicates that the computed Pareto front directly covers the whole reference Pareto front without any distances.
\begin{equation}
    \label{eq:eval:wh:gd}
    GD = \frac{\sqrt{\sum_{i=1}^{|PF_c|} d(s_i, PF_{ref})^2}}{|PF_c|}
\end{equation}

Third, the ED performance indicator defined in \cref{eq:eval:wh:ed} measures the Euclidean distance from a reference solution to its closest solution in the computed Pareto front~($d(s_{ref}, PF_c)$).
The reference solution is defined similar to the reference Pareto front and selects the best known values among all retrieved solutions for each objective and combines them into one solution value.
The best possible value for this metric is zero, indicating that the best solution of the computed Pareto front matches the reference solution.
\begin{equation}
    \label{eq:eval:wh:ed}
    ED = d(s_{ref}, PF_c)
\end{equation}

Fourth, the PFS performance indicator measures the number of solutions in the computed Pareto front as defined in \cref{eq:eval:wh:pfs}.
A larger PFS value indicates a higher diversity of the computed solutions and, hence, the user has more options to choose from. 
\begin{equation}
    \label{eq:eval:wh:pfs}
    PFS(PF_c) = |PF_c|
\end{equation}

Fifth, the GS performance indicator defined in \cref{eq:eval:wh:gs} measures the diversity of the solutions that exist in the computed Pareto front.
Therefore, we calculate a set of extreme solutions from the reference Pareto front called $e_1,\dots,e_m$ where $e_i$ has the best value for the i-th objective function~($f_i$). 
$d(e_k, PF_c)$ refers to the Euclidean distance from extreme solution $e_k$ to its nearest solution in $PF_c$.
$d(s, PF_c)$ calculates the Euclidean distance from the solution $s \in PF_c$  to its nearest solution in $PF_c$ and $\bar{d}$ represents the mean of these distances over all solutions in $PF_c$.
This performance indicator measures how even the solutions in $PF_c$ are spread, where a lower GS value indicates a more even distribution.
\begin{equation}
    \label{eq:eval:wh:gs}
    GS(PF_c,PF_{ref}) = \frac{\sum_{k=1}^m d(e_k, PF_c) + \sum_{s\in PF_c} \left|d(s, PF_c) - \bar{d}\right|}{\sum_{k=1}^m d(e_k, PF_c) + |PF_c| \cdot \bar{d}}
\end{equation}

Sixth, the IGD performance indicator combines convergence and diversity aspects of $PF_{ref}$ and is defined in \cref{eq:eval:wh:igd}.
Again, we use the Euclidean distance of a solution $s_i$ in $PF_{ref}$ to its closest solution in $PF_c$ which we define as $d(s_i, PF_c)$).
We set the sum over all solutions in the reference Pareto front in the ratio to the size of the reference Pareto front.
We use this value to compare two computed Pareto fronts where the computed Pareto front with the lowerIGD value is closer to the reference Pareto front, and hence, we consider it better than the other computed Pareto front.
\begin{equation}
    \label{eq:eval:wh:igd}
    IGD(PF_c, PF_{ref}) = \frac{\sqrt{\sum_{i=1}^{|PF_{ref}|} d(s_i, PF_c)^2}}{|PF_{ref}|}
\end{equation}

Finally, the HV performance indicator defined in \cref{eq:found:qi:hv} also deals with the combination of convergence and diversity aspects of the computed solutions.
It measures the volume in the objective space that the $PF_c$ covers with respect to a given reference point. 
Thus, it computes the volume of the hypercube resulting from the diagonal corners $v_i$ between the solutions $s_i$ in the $PF_c$ and the reference point $P_{ref}$ having the worst objective function values.
The reference point must not overlap with the values of $PF_c$, and can therefore be defined outside the range of values of the objective functions---below the value range of values for a maximization problem and above the range for a minimization problem.
A higher HV value in general indicates a better performance of the $PF_c$.
\begin{equation}
    HV(PF_c, P_{ref}) = volume ( \cup_{i=1}^{PF_c} v_i)
    \label{eq:found:qi:hv}
\end{equation}

\subsection{Optimization Algorithms}
\label{sec:found:optalgo}
Research in the field of optimization algorithms is a very old discipline and, accordingly, has already developed a wide range of techniques.
Following the structure of \cite{rao2019engineering} and \cite{brownlee2011clever}, optimization algorithms can be divided into the following categories: exact algorithms, mathematical programming techniques, stochastic algorithms, physical algorithms, probabilistic algorithms, evolutionary algorithms, swarm algorithms, immune algorithms and neural algorithms.
For a comprehensive overview of the most common algorithms from these categories, we refer the reader to the books~\cite{rao2019engineering,brownlee2011clever}.
In the following, we limit the discussion of optimization algorithms to those used in this thesis.

The first category of optimization algorithms are exact algorithms with the most general brute-force algorithm~\cite{pearl1987search}.
This algorithm does not require any domain-specific knowledge, but operates on a set of states, starting from an initial state and using legal operators.
Breadth-first search or depth-first search are two specific examples of brute-force algorithms.
The brute-force algorithms perform an exhaustive search and explore the entire design space, which is the reason for their exponential time complexity.
However, the advantage of these algorithms is that they are guaranteed to find the best solution.

LS is an example algorithm from the category of stochastic algorithms~\cite{brownlee2011clever,lourencco2001beginner}.
This algorithm starts from an initial constructed solution and explores the direct neighborhood of that solution.
Usually, the neighborhood can be found by replacing one decision variable of the current solution. 
If the neighboring solution has better objective function values, the algorithm uses it as the next starting point. 
If the neighbor solutions do not lead to better values, LS terminates and returns the solution. 
If the LS examines all neighboring solutions and selects the one that maximizes the objective function value, this variant is called Hill Climbing.

From the category of probabilistic algorithms, we consider Bayesian Optimization~\cite{pelikan2000linkage,brownlee2011clever}.
This algorithm builds a probabilistic model of the joint distribution of promising solutions using Bayesian networks. 
At each iteration, the algorithm queries one observation point from the objective function and updates its model accordingly.
An additional acquisition function defines the most promising candidate for the next observation point.
The algorithm terminates after a maximum amount of iterations and returns the best solution found so far. 

Simulated Annealing is one of the physical optimization algorithms~\cite{kirkpatrick1983optimization,brownlee2011clever}.
It was inspired by the physical process of annealing in metallurgy and involves heating and cooling metal to increase the strength and durability of the material. 
The idea is that as the temperature of a material increases, the degrees of freedom within the system also increase and more changes are possible.
As the temperature decreases, the possibility of changes to the system becomes smaller.
The algorithm starts with an initial solution and changes it iteratively.
Therefore, two functions are relevant: a temperature function and the objective function.
As long as the temperature is still high, the objective function has less influence on the selection of the next candidate solution.
With decreasing temperature, the objective function gets more weight and the algorithm only considers solutions with better objective function values.

From the category of evolutionary algorithms, we consider the GA~\cite{holland1992genetic,brownlee2011clever}.
It is inspired by the process of natural selection and the evolution of a population by recombination and mutation of individuals representing a solution encoded as a genome. 
The algorithm starts with an initial set of solutions, called the population, and performs selection, crossover, and mutation in each iteration.
Selection uses the objective function to evaluate the fitness of individuals and select the individuals for recombination.
Then, the crossover procedure combines the genomes of the selected individuals to breed a new individual as part of the offspring.
Each new individual is then mutated to increase the diversity of the population.
Finally, as the population increases in each iteration, the individuals with the lowest fitness value are discarded in each iteration to achieve the predefined population size.
A multi-objective version of the GA is the NSGA-II, which computes a Pareto front and aims at a good distribution of solutions by considering the crowding distance as density estimate of the solutions in the front~\cite{deb2002fast}.

Finally, ACO is a representative of swarm algorithms~\cite{dorigo1996ant,brownlee2011clever}.
ACO is inspired by the behavior of ants in search of food and uses the concept of pheromones that ants leave on a good path between their colony and food sources.
The algorithm first simulates a random movement of ants in the environment.
Once an ant discovers a good food source, it begins emitting pheromones along the path back to the colony.
When other ants notice the pheromone trail, they follow it and increase the concentration of pheromones along the way.
However, some of the ants do not follow the path and keep exploring new paths to find better routes and create new pheromone trails.
As the pheromone in the environment decreases, the shortest path receives the highest pheromone concentration during the optimization process.
After a predefined period of time, the algorithm terminates with the shortest path represented by the pheromone trail with the highest concentration.

\section{Foundations on Mezzanine Warehouses}
Warehouses play a central role in the supply chain of a company and contribute to its logistical success.
When employing humans, picker-to-parts and parts-to-picker methods are differentiated~\cite{Koster2007}.
Experts estimate the picker-to-parts system to be the most common in Western Europe with a share of over~80\%~\cite{Dekoster2007}.
A well-known picker-to-parts system is the mezzanine warehouse which we address in this work.
Working within a mezzanine warehouse consists of two main tasks: (i)~filling the storage with goods~(storage assignment) and (ii)~picking items out of the storage~(order picking).
The storage assignment problem defines the task of selecting storage locations to put a product into storage.
The order picking problem defines the task of computing a pick route that collects the requested products of a customer order.
Finding suitable storage allocations is important, as the allocation of products affects the travel distances during order picking.
Due to the NP-hardness and, hence, the complexity of the storage assignment and the order picking problem, efficient optimization algorithms are required to find satisfying solutions within acceptable times.
This section first introduces the mezzanine warehouse layout in Section~\ref{sec:found:wh:mezzanine} and presents state-of-the-art mechanisms for storage assignment and order picking in Section~\ref{sec:found:wh:sa} and Section~\ref{sec:found:wh:op}, respectively.

\subsection{Warehouse Layout}
\label{sec:found:wh:mezzanine}
\begin{figure*}[htb]
	\centering
	\includegraphics[width=0.7\textwidth]{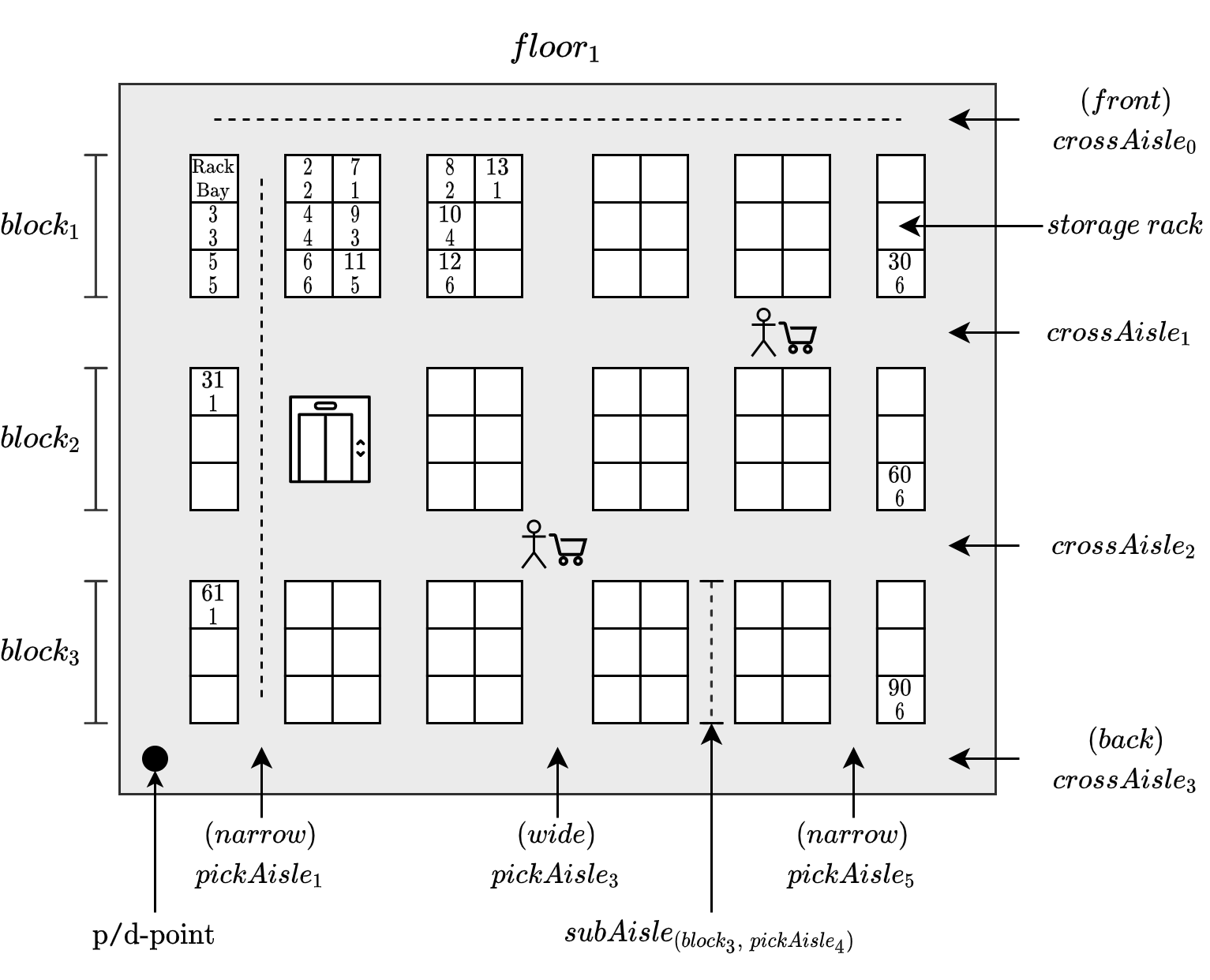}
	\caption[Example floor layout of a mezzanine warehouse.]{Example floor layout of a mezzanine warehouse (c.f.~\cite{Roodbergen2001}) consisting of blocks, storage racks, cross and pick aisles, and p/d-points.}
	\label{fig:found:wh_layout}
\end{figure*}

Mezzanine warehouses usually store small-sized products that need to be picked by employees traveling through the warehouse.
This type of warehouse consists of one or multiple floors to store goods using racks. 
Roodbergen and De Koster~\cite{Roodbergen2001} provide a general layout of such a mezzanine warehouse floor in their work from top-down view from which the following illustration in~\ref{fig:found:wh_layout} is derived.

Each floor consists of a predefined number of storage racks illustrated as white squares arranged in blocks, each consisting of three storage racks.
The racks can be identified by their rack id depicted as the top number inside the rack, while the bottom number indicates the bay number.
While the rack id uniquely identifies a rack within a floor, the bay number indicates the ordering of the racks within each block and, hence, is unique solely within a block.
The blocks are separated by aisles of different sizes and directions: (i) horizontal cross aisles and (ii) pick aisles. 
While the horizontal cross aisles do not provide access to the racks, these are used to change the pick aisles from which the employee can access the racks.
The cross aisles are wide enough to travel using picking carts.
The pick aisles can be grouped into two types: narrow and wide pick aisles.
Picking carts can be carried only in wide pick aisles and the employee needs to park the cart at a wide aisle to pick goods within narrow pick aisles.
The part of an aisle within a block is called sub aisle of the according block, and, hence, a pick aisle consists of multiple sub aisles.
The black dot at the left bottom of the figure indicates a p/d-point where employees need to deliver the picked goods or pickup the next set of goods to be stored in the warehouse.
Finally, the employees are able to change the floor of the warehouse by using stairs or lifts.

\begin{figure*}[htb]
	\centering
	\includegraphics[width=0.7\textwidth]{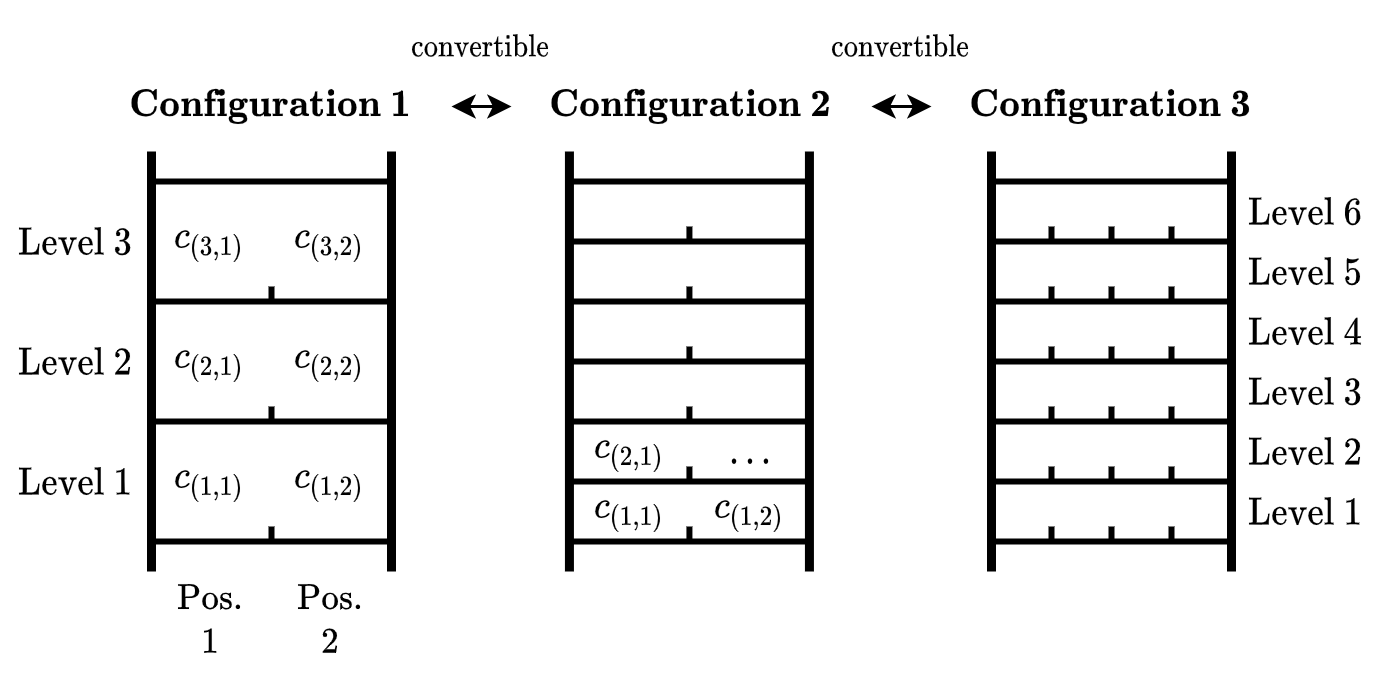}
	\caption{Example rack configurations that can be individualized with respect to the requirements of the stored goods.}
	\label{fig:found:wh_racks}
\end{figure*}

The considered racks are identical in terms of their height, width, and depth within the warehouse.
However, each rack can be configured individually with respect to the needs of the currently stored goods.
Figure~\ref{fig:found:wh_racks} shows possible configurations of the considered racks in this work depicted from the front.
The rack configuration determines the number of shelves, that is, number of levels within a rack, and the number of compartments per shelf.
Configuration~1 in the figure depicts three shelves each divided into two parts that represent compartments. 
Hence, this configuration offers six storage locations.
Configuration~2 offers twelve storage locations by applying six shelves with two compartments each, and Configuration~3 offers 24 storage locations by dividing the six shelves into four compartments each.

\subsection{Storage Assignment}
\label{sec:found:wh:sa}
The storage assignment problem defines the task of selecting storage locations to put a product into storage.
Since mezzanine warehouses usually provide a large number of storage racks, it is difficult to find the optimum storage allocation that fulfills all custom constraints as well as ergonomic and economic objectives defined for the problem.
The following sections introduce a subset of the most common storage assignment strategies present in the literature and applied in real-world warehouses.

The simplest storage policy is called \textit{dedicated storage policy}~\cite{Bartholdi2019} in which each product is assigned to a dedicated and exclusive storage location.
Using this policy, no changes in the warehouse need to be made and employees get to know the locations over time.
However, the warehouse utilization is comparably low with a value of half of the storage capacity on average.

The \textit{random storage policy}~\cite{Bartholdi2019} does not exclusively reserve locations for specific products but assigns incoming products randomly to unoccupied racks. 
This policy usually achieves better utilization values but comes with a higher administrative effort since the product locations change over time.

The \textit{closest open location storage policy}~\cite{Dekoster2007} reduces the randomness of the random storage policy by letting the employees select the storage location which results in selecting the first empty location the employee encounters.
This leads to a higher utilization in the neighborhood of the p/d-points while racks that are farther tend to be empty.

The \textit{rank-based storage policy} removes the randomness completely and ranks each incoming product based on a predefined rule set.
These rules could contain but are not limited to~\cite{Petersen2005,frazelle2002supply,Heskett1963Cube}: popularity, turnover, that is, the requested quantity by customers, the volume, the pick density, or the cube-per-order index.

\begin{figure*}[htb]
	\centering
	\includegraphics[width=0.7\textwidth]{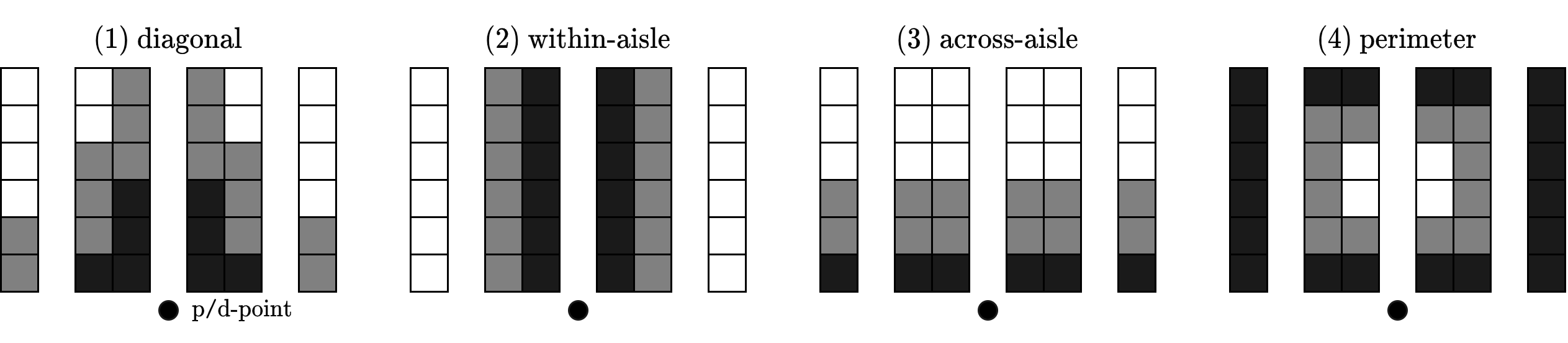}
	\caption{Illustration of rank-based storage assignment strategies (c.f.~\cite{Petersen1999Evaluation}). }
	\label{fig:found:wh:sa_policies}
\end{figure*}

Figure~\ref{fig:found:wh:sa_policies} illustrates four additional rank-based storage assignment policies introduced by~\cite{Petersen1999Evaluation}.
This policy assigns storage locations based on the best- and worst-ranked products illustrated by black and white squares in the figure determined by their distance to the next p/d-point.
The diagonal strategy~(1) assigns incoming products according to their Euclidean distance to the next p/d-point, best ranked close and worst ranked further away from the p/d-point.
The within-aisle strategy~(2) assigns the best-ranked incoming products within the same aisle of the p/d-point.
The across-aisle strategy~(3) assigns the best-ranked incoming products to the entrance of all aisles that is nearest to the p/d-point.
Finally, the perimeter strategy~(4) assigns the best-ranked incoming products around the perimeter of the warehouse assuming that these are the most traveled aisles of the floor.

\textit{Class-based storage assignment strategies}~\cite{Dekoster2007} are a combination of the previously mentioned strategies as they include the following three tasks: (i)~grouping of products into classes, (ii)~definition of class regions within the warehouse, (iii)~assign products to the defined region.
Usually, the grouping is done using three classes~(A-, B-, and C-class) that distinguish between fast- and slow-moving products.

Further, De~Koster et al.~\cite{Dekoster2007} take another factor into account for determining the optimal storage locations for products in their \textit{family grouping strategy}.
They include the correlation of products into account and propose to store products close to each other that often need to be picked in combination.
They differentiate two types of strategies: complimentary-based and contact-based, which measure joint demand or contact frequencies of the correlated products, respectively.

Besides the assignment of products to racks, the \textit{golden zone assignment strategies}~\cite{Petersen2005} focus on the assignment of products into compartments.
With golden zone these strategies refer to compartments located at grip height, that is between waist and shoulders of the picker.
These strategies assign fast-moving products exclusively to the golden zones of racks and disregard the travel distance to the according rack.

\subsection{Order Picking}
\label{sec:found:wh:op}
The order picking problem defines the task of constructing pick routes within a warehouse that include all products of a pick list derived by a customer order.
State-of-the-art routing heuristics are able to construct these routes fast and try to minimize the travel distance per route. 
They model the problem as TSP and start and end the route at a specific p/d-point.

\begin{figure*}[htb]
	\centering
	\includegraphics[width=0.7\textwidth]{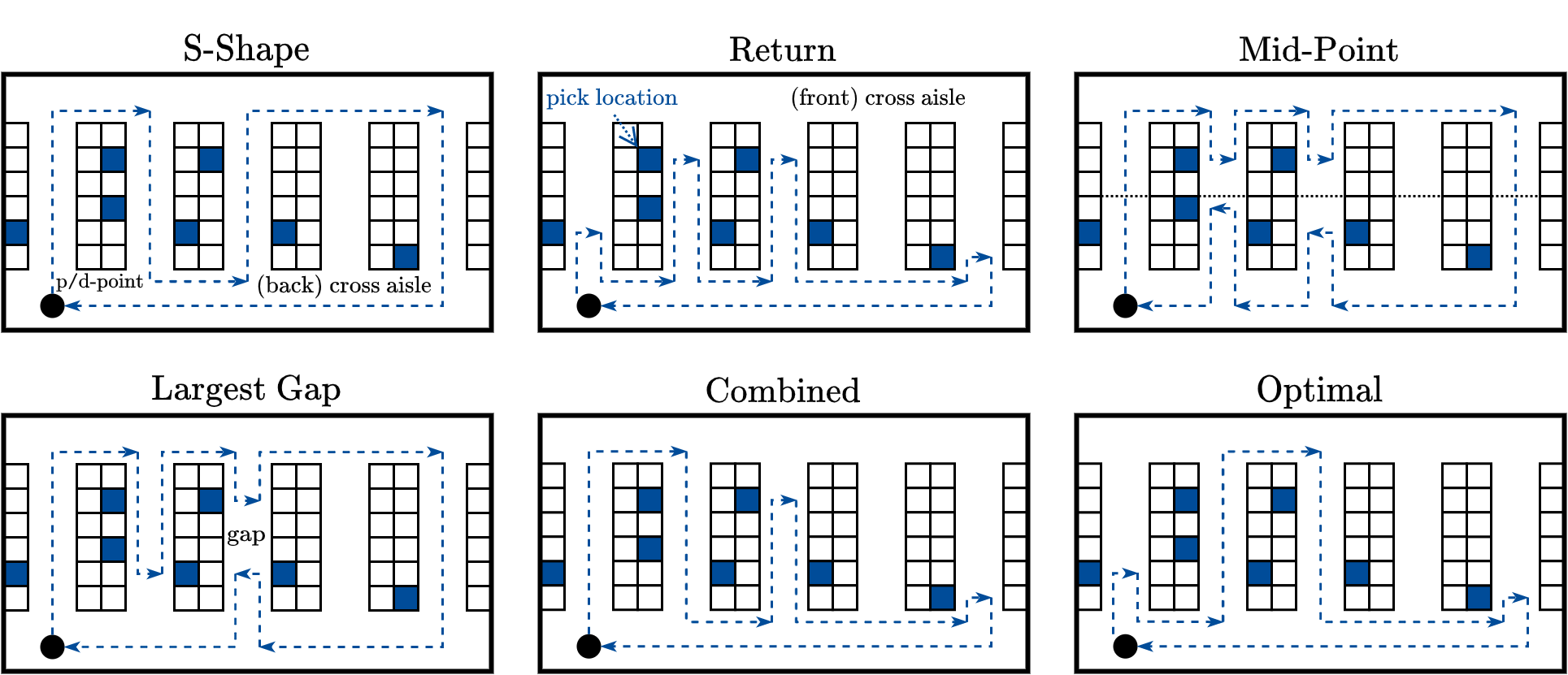}
	\caption{Illustration of five state-of-the-art order picking strategies and the optimal strategy based on dynamic programming (c.f.~\cite{Petersen1997,Ratliff1983}).}
	\label{fig:found:wh:op_policies}
\end{figure*}

Petersen~\cite{Petersen1997} proposes five routing strategies applicable for mezzanine warehouses and which are illustrated in Figure~\ref{fig:found:wh:op_policies}.
The \textit{S-Shape} strategy defines the route inside the warehouse to completely traverse all aisles that contain required products.
It alternates the traversing direction so that a shape similar to the letter S is created as depicted in the figure.
When applying the \textit{Return} strategy, the picker enters the pick aisles in which required products are stored but always returns to the entrance of this aisle.
Hence, in the worst case, an aisle might be traversed two times in case the product to be picked is stored in the last rack.
In the \textit{Mid-Point} strategy, the picker passes through each aisle at most to the middle of the aisle and then returns to the entrance through which he entered the aisle.
In case a product is located further within the aisle, that is, behind the mid-point, the picker needs to enter the pick aisle from the other entrance again. 
The \textit{Largest Gap} strategy adapts the idea of the Mid-Point strategy but dynamically sets the point that should not be traversed based on the largest gap between two products in the aisle.
In this way, this strategy attempts to minimize travel distance by avoiding passing shelves that are not needed.
Finally, the \textit{Combined} strategy combines the S-Shape and Return strategies.
It selects the pick aisle entry based on the current location of the picker and after all products are picked, the strategy decides to either complete this aisle and use the other entrance or to return to the initial entrance.
In addition to these five strategies, Ratliff and Rosenthal~\cite{Ratliff1983} present another strategy called \textit{Optimal} strategy where they apply dynamic programming to find the shortest route.
For small problem instances, this method can be used to determine the optimum solution that can be used as gold-standard.

\section{Meta-Model of Considered Mezzanine Warehouses}
\label{sec:metamodel}
\begin{figure*}[htb]
\centering
\includegraphics[width=0.7\textwidth]{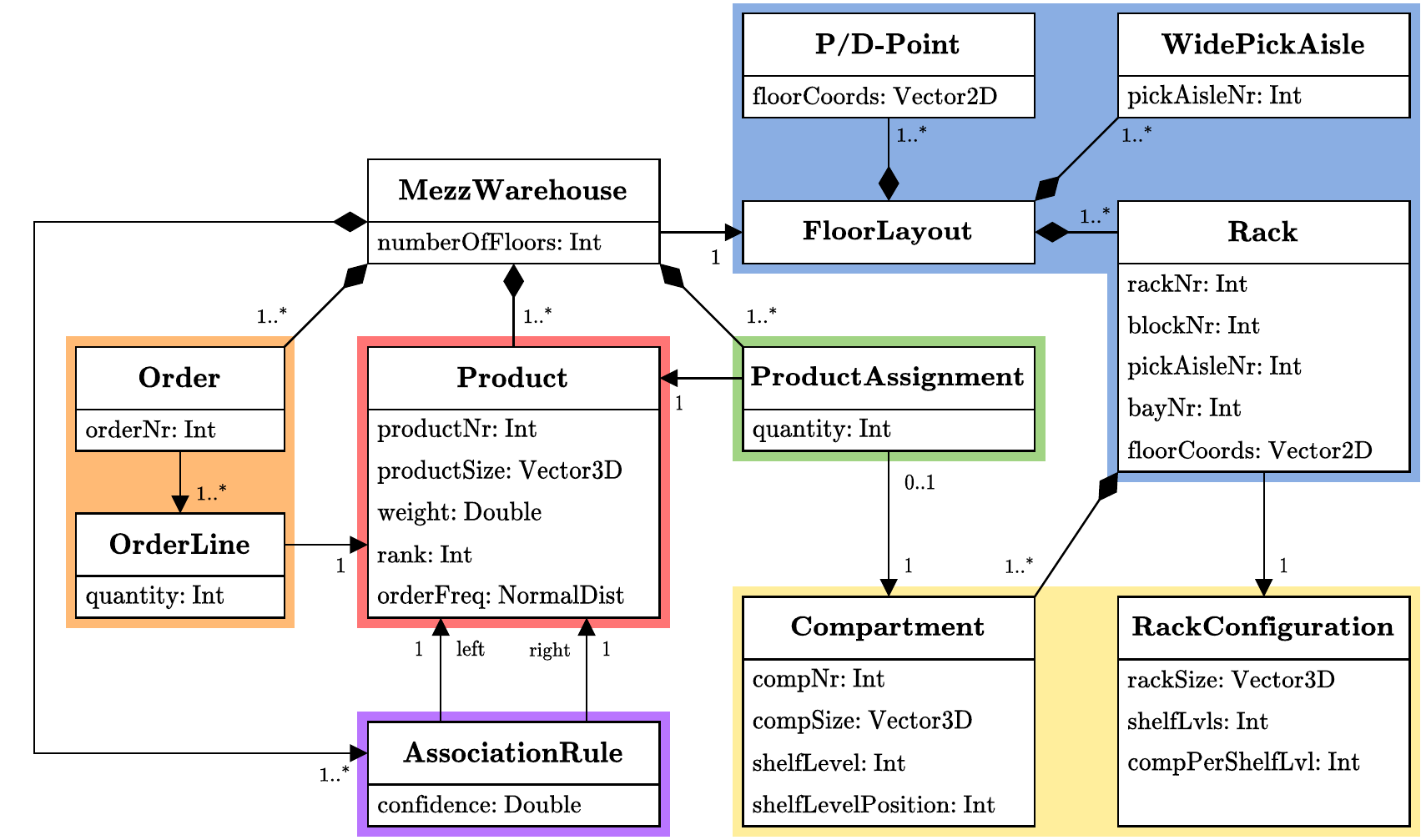}
\caption{The meta-model describes the structure and state of mezzanine warehouses.}
\label{fig:metamodel}
\end{figure*}

The storage assignment and order picking algorithm require information on the warehouse layout, the product assortment, the products' storage locations, and the current state of the warehouse.
Figure~\ref{fig:metamodel} illustrates our proposed meta-model.

\begin{figure*}[tb]
\centering
\includegraphics[width=0.7\textwidth]{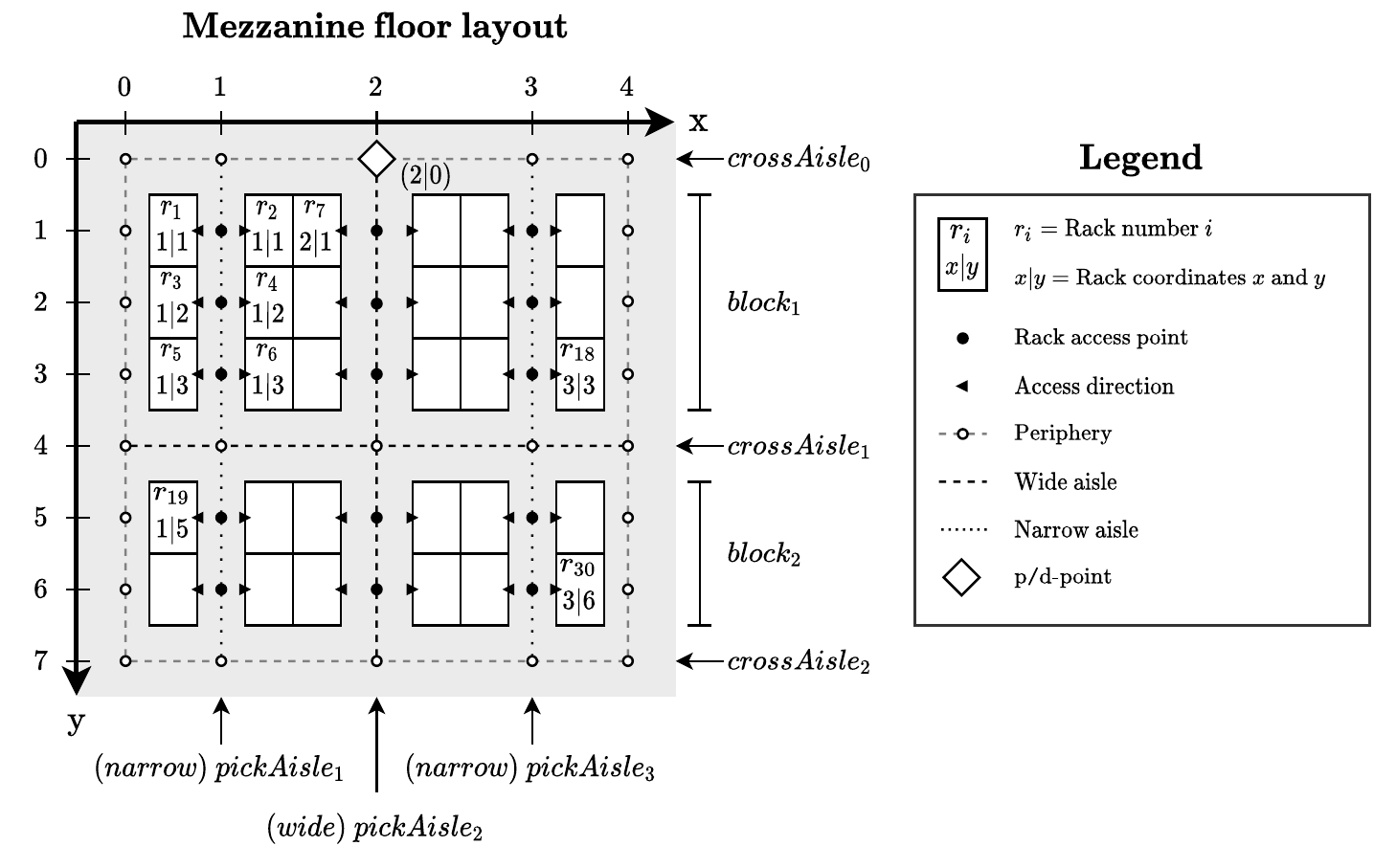}
\caption{Example mezzanine floor layout from top-down view.}
\label{fig:layout}
\end{figure*}

The blue box describes the floor layout defining the arrangement of racks within one floor of the mezzanine warehouse~(\texttt{FloorLayout}). 
Each floor consists of the classes, \texttt{P/D-Point}, \texttt{WidePickAisle}, and \texttt{Rack}.
A p\texttt{/}d-point is the pickup and delivery point where personal needs to collect items to be stored in the warehouse or deliver items of a customer order that were collected.
Regard the class \texttt{WidePickAisle}, two types of pick aisles exist: wide and narrow pick aisles.
In wide pick aisles, pickers can take along their pick cart to cross the aisle while it needs to be parked at the aisle entry for narrow pick aisles.
A floor can be illustrated as a two-dimensional map as depicted in
Figure~\ref{fig:layout}:
The racks with their unique identifiers $r_3$ and $r_4$ are assigned the floor coordinates $x=1$ and $y=2$ since their access points are both located at $(1|2)$.
The vertical aisles located at $x=0$ and $x=4$, as well as the horizontal cross aisles at $y=0$, $y=4$, and $y=7$, form the periphery of the floor.
Periphery aisles usually contain the p\texttt{/}d-points~(e.g. at $(2|0)$).
A wide pick aisle is depicted at x-coordinate two and two narrow pick aisles are shown at x-coordinates one and three, where the picker needs to park his pick cart.
Real-world mezzanine warehouses may apply different layouts on each floor; however, we assume that each floor in the mezzanine warehouse applies the same layout.

Since diagonal movements are not possible in this layout, the Manhattan distance function is applied to calculate the distance between two locations $p$ and~$q$:
\begin{equation}
\label{eq:distance}
distance(p,q) = \sum_{i=1}^n |p_i - q_i|
\end{equation}

The classes inside the yellow box~(\texttt{Compartment} and \texttt{RackConfiguration}) define the configuration of a rack, referring to its size, the number of shelf levels, and the number of compartments per shelf level.
The \texttt{Compartment} class includes an identifier and a three-dimensional vector specifying the compartment's dimensions.
The shelf level and the shelf level position defines the compartment's location within the rack.
The class \texttt{Product} defines the products using five properties: product number, size, weight, rank, and order frequency.
The rank ($\geq$ 1) allows identifying fast and slow-moving products by the frequency at which the product appears in recent customer orders.
The product of rank 1 represents the most frequently ordered product.
The order frequency describes the frequency to which a product is usually ordered using a gaussian distribution.
Both properties are derived from recent customer orders and represent redundant information which prevents the algorithms from recalculating this information each time they need it.
Further, these properties are later used in the storage assignment optimization to find better racks regarding their frequency and usual ordered amount.
The class \texttt{ProductAssignment} specifies the quantity of which a product is assigned to a specific compartment.
The classes \texttt{Order} and \texttt{OrderLine} of the orange package define the structure of a customer order consisting of a unique order number and multiple order lines.
An order line specifies the quantity to which a product is ordered.
The class \texttt{AssociationRule} defines association rules derived by the Apriori algorithm~\cite{injectingProblemDependentKnowledge}.
The confidence ranges from 0 to 1 and expresses the strength of the correlation between the left-sided and the right-sided set of products.
These rules are used in the storage assignment algorithm later on to store correlated products close to each other which may increase the order picking performance. 

%----------------------------------------------------
%----------- Storage Assignment Algorithm -----------
%----------------------------------------------------
\section{Storage Assignment}
\label{sec:storageassignment}
The overall goal of the storage assignment algorithm is to select a set of compartments for storing an incoming product by considering multiple economic and ergonomic constraints simultaneously.

\subsection{Constraints and Assumptions}
In expert interviews, we identified multiple hard constraints that should be covered in our approaches.
These hard constraints specify whether a storage allocation is considered feasible and a feasible solution never violates any of these constraints:
Each incoming item must be assigned to a compartment~($HC_1$). 
The selected compartment must either be empty or partially occupied by items of the same product~($HC_2$).
Each item has to fit in the remaining free space if its compartment~($HC_3$).
Furthermore, we define multiple soft constraints that measure the extent to which a storage allocation fulfills economic criteria:
The products should be evenly spread on each floor~($SC_1$).
Fast-moving products should be assigned close to a p/d-point~($SC_2$).
The mean ordered quantity of a product should be locally available~($SC_3$).
Correlated products should be stored close to each other~($SC_4$).
The storage space should be used as efficiently as possible~($SC_5$).
Finally, we define two ergonomic soft constraints:
Heavy products should be stored at grip height~($SC_6$).
Fast-moving products should be assigned to compartments at grip height~($SC_7$).

Further, we state the following assumptions for our approach:
The state of the warehouse does not change while the storage assignment algorithm is running.
Thus, the products are not repositioned nor removed, and the racks' configurations do not change.
Further, the algorithm allocates only one product at a time.
The storage racks may apply different rack configurations and products may only be assigned to fitting compartments.
A compartment is allowed to store multiple items of the same product but may not store two different products at the same time. 
Finally, product ranks and association rules are derived from recent customer orders.

\subsection{3-Phase Storage Assignment Algorithm}
Our storage assignment algorithm consists of three phases that intend to reduce the complexity of the optimization problem: 
(i)~assignment of products to floors, (ii)~assignment to racks w.r.t. economic criteria, and (iii)~assignment to compartments w.r.t. ergonomic criteria.

In the first phase, the incoming product quantity is split among the mezzanine floors~($SC_1$) so that each floor provides the same quantity of the product.
This way, we try to reduce the required floor changes during a pick route to a minimum.
Thus, we first determine the total quantity of the incoming product that is already available in each floor, calculate the ideal quantity for each floor after storage assignment, and assign the missing quantity to each floor.
Remaining items, due to rounded results, are allocated to a random floor.

The second phase addresses the economic soft constraints $SC_2$ to $SC_5$ and aims to reduce travel distances during order picking.
This phase assigns the incoming products to racks on a specific floor.
Since this phase requires optimizing a set of constraints, we apply a multi-objective optimization algorithm that is described in Section~\ref{sec:sa_phase2}.

The third phase aims to satisfy the ergonomic soft constraints $SC_6$ and $SC_7$.
We classify a product $p$ into three weight classes: \textit{light} (up to 3\,kg), \textit{medium} (between 3\,kg and 7\,kg), and \textit{heavy} (over 7\,kg).
We set the grip height to be between 0.75\,m to 1.25\,m and refer to compartments below/above the grip height as low/high zone compartments.
Additionally, we distinguish \textit{fast-moving}, \textit{moderately-moving}, and \textit{slow-moving} products by their relative rank.
The relative rank of a product $p$ calculates as $rank_p / |P|$, where $rank_p$ denotes the rank of product $p$, and $|P|$ the size of the product assortment.
In the first step, the incoming items are assigned to the rack's compartments that already provide items of the same product.
In the second step, the remaining incoming items are assigned to the rack's unoccupied compartments based on predefined penalty values.
The penalty values range from zero to three and the more a compartment~$c$ is unsuited for storing the product $p$, the more penalty points are given~(see~Tables~\ref{tab:penaltiesWeight} and~\ref{tab:penaltiesRank}).

\begin{table}[htb]
    \centering
    \caption{Penalties for assigning a product to a specific compartment with regards to the product weight.}
	\label{tab:penaltiesWeight}
    \begin{tabular}{l l l}
         \toprule
			Zone & Weight & Penalty \\\midrule
			high & light & 0 \\
			high & medium & 2 \\
			high & heavy & 3 \\
			grip height & light & 1 \\
			grip height & medium & 0 \\
			grip height & heavy & 0 \\
			low & light & 0\\
			low & medium & 1 \\
			low & heavy & 1 \\\bottomrule
    \end{tabular}
\end{table}

\begin{table}[htb]
	\centering
	\caption{Penalties for assigning a product to a specific compartment with regards to the product rank.}
	\label{tab:penaltiesRank}
	\begin{tabular}{l l l}
		\toprule
		Zone & Rank & Penalty \\\midrule
		high & slow & 0 \\
		high & moderate & 0 \\
		high & fast & 2 \\
		grip height & slow & 3 \\
		grip height & moderate & 1 \\
		grip height & fast & 0 \\
		low & slow & 0 \\
		low & moderate & 0 \\
		low & fast & 2 \\	\bottomrule
	\end{tabular}
\end{table}

\section{Genetic Algorithm for Storage Assignment}
\label{sec:sa_phase2}
This section presents our custom version of the NSGA-II algorithm that was proposed by~\cite{Deb2002}. 
The algorithm receives the current state of a floor and assigns the incoming items to a set of racks on this floor.
Note that the NSGA-II is executed for each floor individually.

\subsection{Chromosome Encoding}
\begin{figure*}[htb]
\centering
\includegraphics[width=0.7\textwidth]{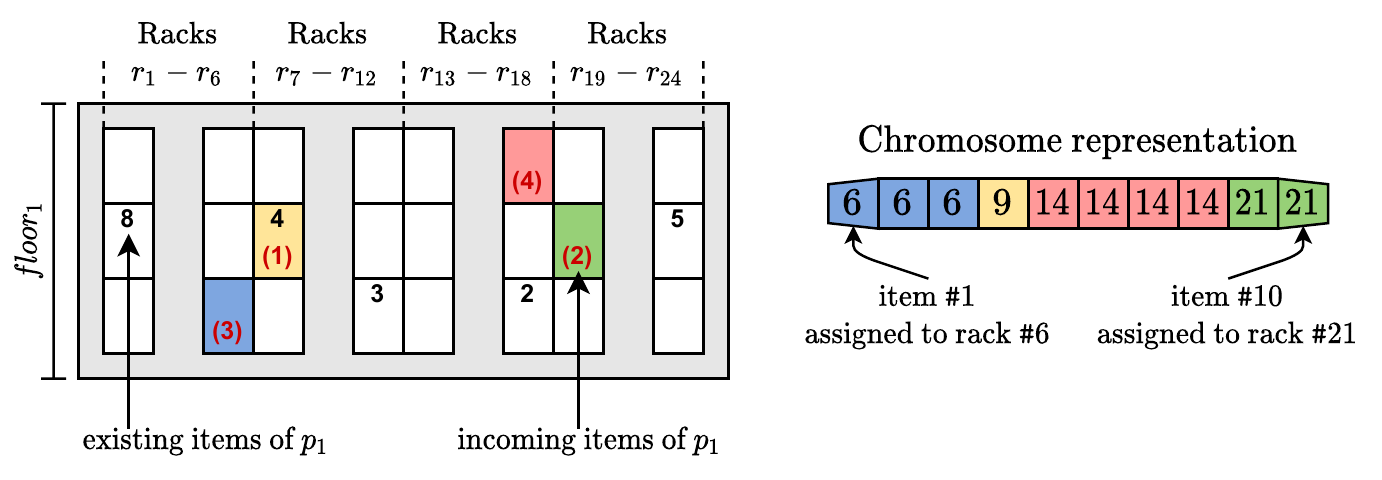}
\caption{A chromosome encodes the racks selected for storing the incoming items.}
\label{fig:chromosomeEncoding}
\end{figure*}

Since the NSGA-II algorithm is a genetic algorithm, we propose the chromosome encoding depicted in Figure~\ref{fig:chromosomeEncoding}.
The figure illustrates an example allocation task where ten items of product $p_1$ must be assigned to the racks on $floor_1$.
The black numbers indicate the existing items of product~$p_1$, while the red numbers indicate the incoming items of product~$p_1$.
The right side shows the chromosome that encodes the storage allocation depicted on the left side by specifying the racks selected for storing each incoming item.
Since ten items of product~$p_1$ are assigned, the chromosome's length equals~10.

\subsection{Objective Functions}
\label{sec:objFunct}
A set of objective functions guide the NSGA-II algorithm to find good storage allocations.
We propose four domain specific objective functions for our maximization problem: (i)~spread score, (ii)~distance score, (iii)~quantity score, (iv)~correlation score.

\subsubsection{Spread Score}
\label{sec:spreadScore}
This score addresses constraint~$SC_1$ and aims to equally spread the incoming quality of product~$p$ across the entire floor.
Hence, we divide the floor~$f_j$ into multiple areas~$A$ of equal size.
To calculate the spread score, we use the total~($totalQ$) and ideal quantity~($idealQ$) of a product in an area of a floor.
The $totalQ$ is the sum of the existing and incoming items in an area, while the $idealQ$ is calculated by dividing the sum of the existing and incoming quantity of product $p$ on the floor by the number of defined areas.
The final spread score for chromosome~$C$ is calculated as the sum of differences between the total and the ideal quantity for all areas~(see Equation~\ref{eq:spreadscore}).
\begin{equation}
\label{eq:spreadscore}
spreadScore_{p, f_j, C}= (-1) \sum_{o=1}^A | idealQ_{p, f_j, j} - totalQ_{p, f_j, j} |
\end{equation}

\subsubsection{Distance Score}
\label{sec:distScore}
\begin{figure*}[htb]
\centering
\includegraphics[width=0.7\textwidth]{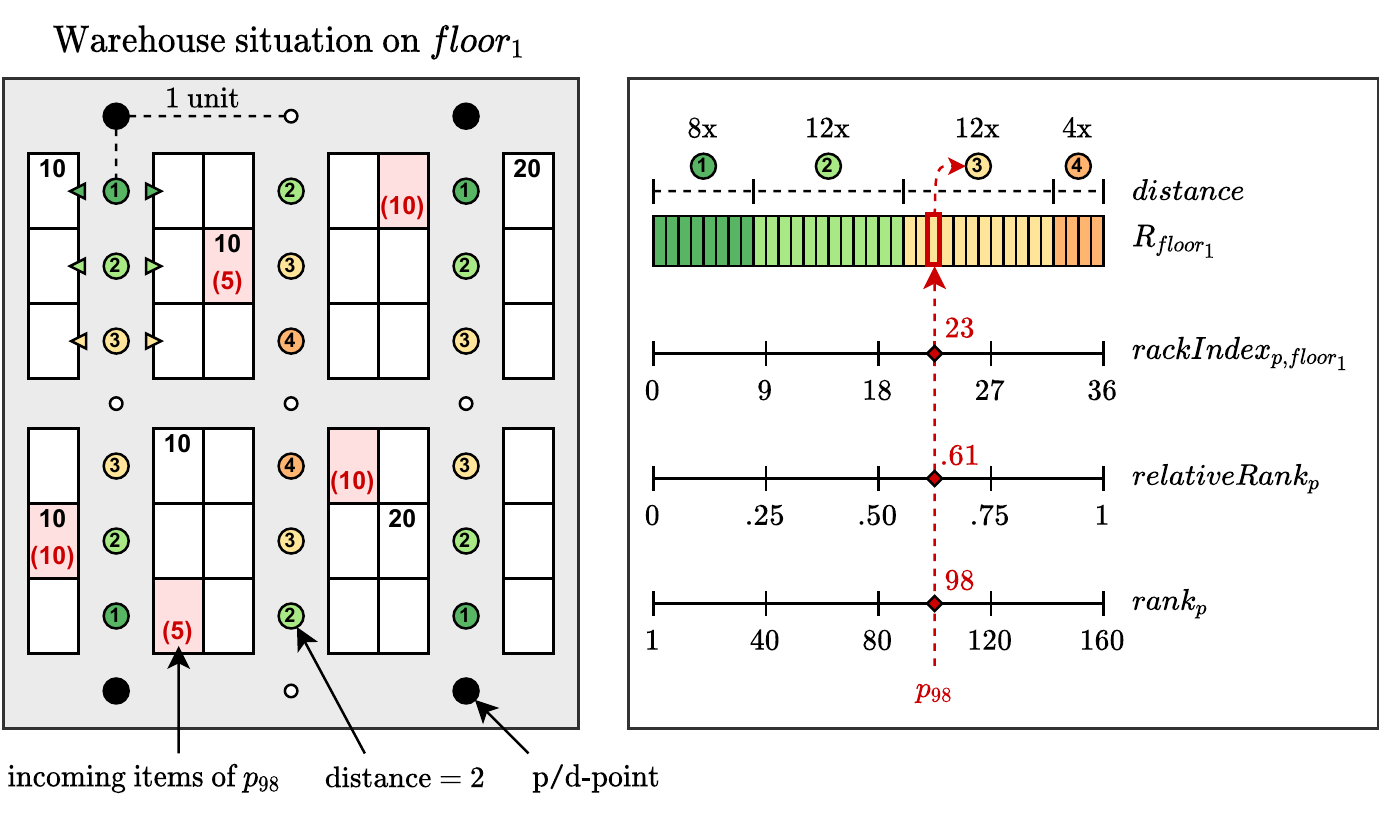}
\caption{Calculating the ideal distance for storing the incoming product $p_{98}$.}
\label{fig:distscore}
\end{figure*}

This score addresses constraint $SC_2$ and aims to allocate slow-moving products to racks further away from the p/d-points.
Hence, the distance score quantifies the extent to which the walking distances~($dist_{r_i}$) of the selected racks match the ideal distance~($idealDist_{p, f_j}$).
For calculating the $idealDist$, we perform the following steps:
First, we determine the relative rank of the incoming product $p$ by dividing the rank of the product~($rank_p$) by the size of the product assortment $P$: $relRank_p = rank_p / |P|$.
Then, the relative rank is mapped to a rack index: $rackIdx_{p,f_j} = relRank_p \cdot |R_{f_j}|$ with $R_{f_j}$ being the list of racks of floor $f_j$ sorted by the racks' walking distances to their closest p/d-point.
The rack in $R_{f_j}$ at index $rackIdx$ represents the best-suited rack for storing product $p$ with regard to constraint $SC_2$.
Finally, the $idealDist$ computes as: $idealDist_{p,f_j} = R_{f_j}[rackIdx_{p,f_j}].distance$.
The overall distance score calculates as the sum over all racks in Chromosome~($C$) of differences between the walking distances of the racks selected for storing product $p$ and the $idealDist$~(see Equation~\ref{eq:distanceScore}).
\begin{equation}
\label{eq:distanceScore}
distanceScore_{p, f_j, C}= (-1) \sum_{n \in C} | idealDist_{p, f_j} - dist_{r_i} |
\end{equation}
Further, we provide an example of this calculation in Figure~\ref{fig:distscore}.

\subsubsection{Quantity Score}
\label{sec:quantScore}
This score assesses $SC_3$ and ensures that the mean ordered quantity of a product is locally available.
Therefore, the target quantity defines the quantity to which the product $p$ should be locally available based on a set of recent customer orders: $tq_p = \lceil \mu_p + 2 \sigma_p \rceil$.
Further, we define four masks and a modifier for each mask to measure the density to which the $tq_p$ is locally available:
$M_1$~equals the size of a rack ($maskMod = 1$), $M_2$~equals the size of two facing racks ($maskMod = 0.75$), $M_3$~is a sliding window with half the sub aisle's length ($maskMod = 0.5$), and $M_4$~covers an entire sub aisle ($maskMod = 0.25$).
Using these masks, we calculate a quantity factor for each sub aisle~($sa$) of a floor and each mask~($M_k$).
Therefore, we select the quantity ($q$) of products inside a mask divided by the target quantity: $qFactor_{p,f_j,sa_l} = max( q_{p,f_j,sa_l}(M_k) / tq_p)$.
This results in a value of 1 if the target quantity is met and a value of 0 if no products can be found within this mask.
This quantity factor is then multiplied by the $maskMod$ to calculate the mask score: $maskScore_{p,f_j,sa_l}(M_k) = maskMod_{M_k} \cdot qFactor_{p,f_j,sa_l}(M_k)$.
The highest possible mask score is 1, indicating that the target quantity is available in a single rack of the sub aisle.
Based on these mask scores, the maximum value is selected to assign a score to each sub aisle: $subAisleScore_{p,f_j,sa_l} = max_{k=1}^4 maskScore_{p,f_j,sa_l}(M_k)$.
The final quantity score computes as the sum of all $subAisleScore$s:
\begin{equation}
\label{eq:objFn3-6}
quantityScore_{p, f_j, C}= \sum_{o=1}^{|SA|}~subAisleScore_{p, f_j, sa_l}
\end{equation}

\begin{figure*}[htb]
\centering
\includegraphics[width=0.7\textwidth]{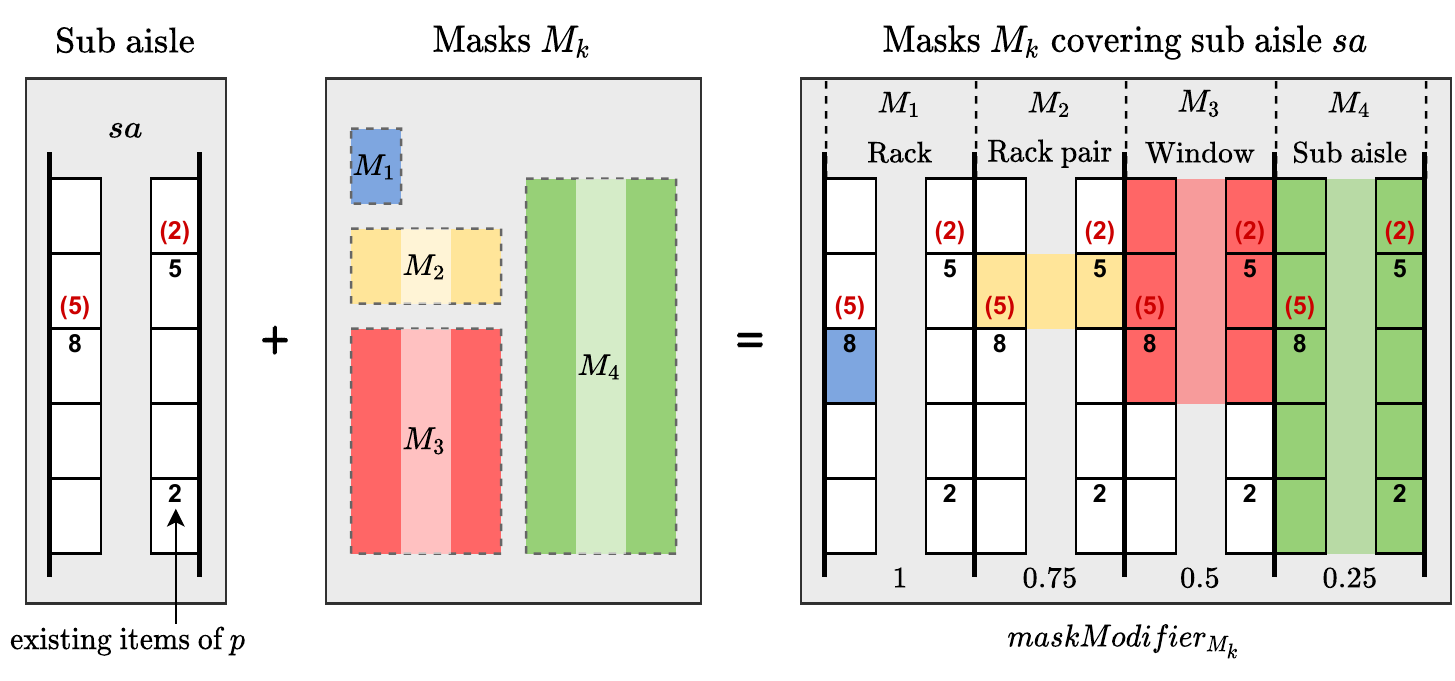}
\caption{The masks $M_k$ count the quantities of product $p$ in the covered regions.}
\label{fig:quantScore}
\end{figure*}

Figure~\ref{fig:quantScore} illustrates the idea of using \textit{masks} of different sizes to measure the density to which the target quantity~$tq_p$ of product $p$ is locally available. 
The left side shows the storage locations of existing and incoming items of product~$p$ in a specific sub aisle~$sa$.
The center of the figure depicts the four masks $M_k$ that iterate over the racks of the sub aisle.
During this process, the masks count the existing and incoming quantities of product~$p$ that can be found in the covered regions.
The right side shows the regions where the masks find the largest quantity of product~$p$ in the sub aisle~$sa$.

\subsubsection{Correlation Score}
This score relates to $SC_4$ and describes the extent to which the incoming product is stored close to its correlated products.
Association rules describe correlations between products and can be derived from recent customer orders.
We consider association rules of the form $rule = \{p\} \xrightarrow{conf} \{cp\}$, where $p$ denotes the incoming product, $cp$ the correlated product, and $conf$ a confidence value.
We first calculate the number of possible clusters of target quantities of the incoming product: $qClusters_{p,f_j} = \lfloor totalQ_{p,f_j} / tq_p \rfloor$.
We use this value to define the ideal quantity to which the correlated product should be available in the vicinity of the incoming product: $idealCorrQ_{rule,f_j} = \lceil qClusters_{p,f_j} \cdot tq_{cp} \cdot conf(rule)\rceil$.
In the next step, we determine the quantity of $cp$ that already is available in the vicinity of $p$.
For this task, the previously introduced masks $M_k$ are used and are placed directly on top of the racks containing $cp$.
Again, the $qFactor$ is calculated to capture the extent to which the target quantity of $p$ is available in the region covered by $M_k$ placed on top of rack $r$: $qFactor_{p, r}(M_k)= q_{p, r}(M_k) / tq_p$. 
Then, we calculate the fraction to which the items of $cp$ stored in $r$ are considered to be in the vicinity of $p$: $corrQ_{rule, r}(M_k)= exQ_{cp, r} \cdot qFactor_{p, r}(M_k) \cdot maskMod_{M_k}$.
$exQ_{cp, r}$ refers to the existing quantity of the correlated product~$cp$ in rack~$r$.
Afterward, we select the $corrQ$ with the highest value representing the mask with the largest amount of $p$ in the vicinity of $cp$:  $corrQ_{rule, r}= \max_{k=1}^{4}~corrQ_{rule, r}(M_k)$.
The sum of all $corrQ_{rule,r}$ over all racks on this floor denotes the quantity of the $cp$ on this floor that is considered as being in the vicinity of $p$: $corrQ_{rule, f_j}= \sum_{rack \in R_{f_j, cp}}~corrQ_{rule, r}$.
Now, we calculated the quantity of the correlated product that is in the vicinity of the incoming product and the difference of this value to the ideal quantity.
Based on this difference, the correlation score is calculated as:
\begin{equation}
\label{eq:objFn4-5}
cS_{p, f_j, C} =  (-1) \sum_{rule \in A_p} idealCorrQ_{rule, f_j} - corrQ_{rule, f_j}
\end{equation}

\begin{figure*}[htb]
\centering
\includegraphics[width=0.7\textwidth]{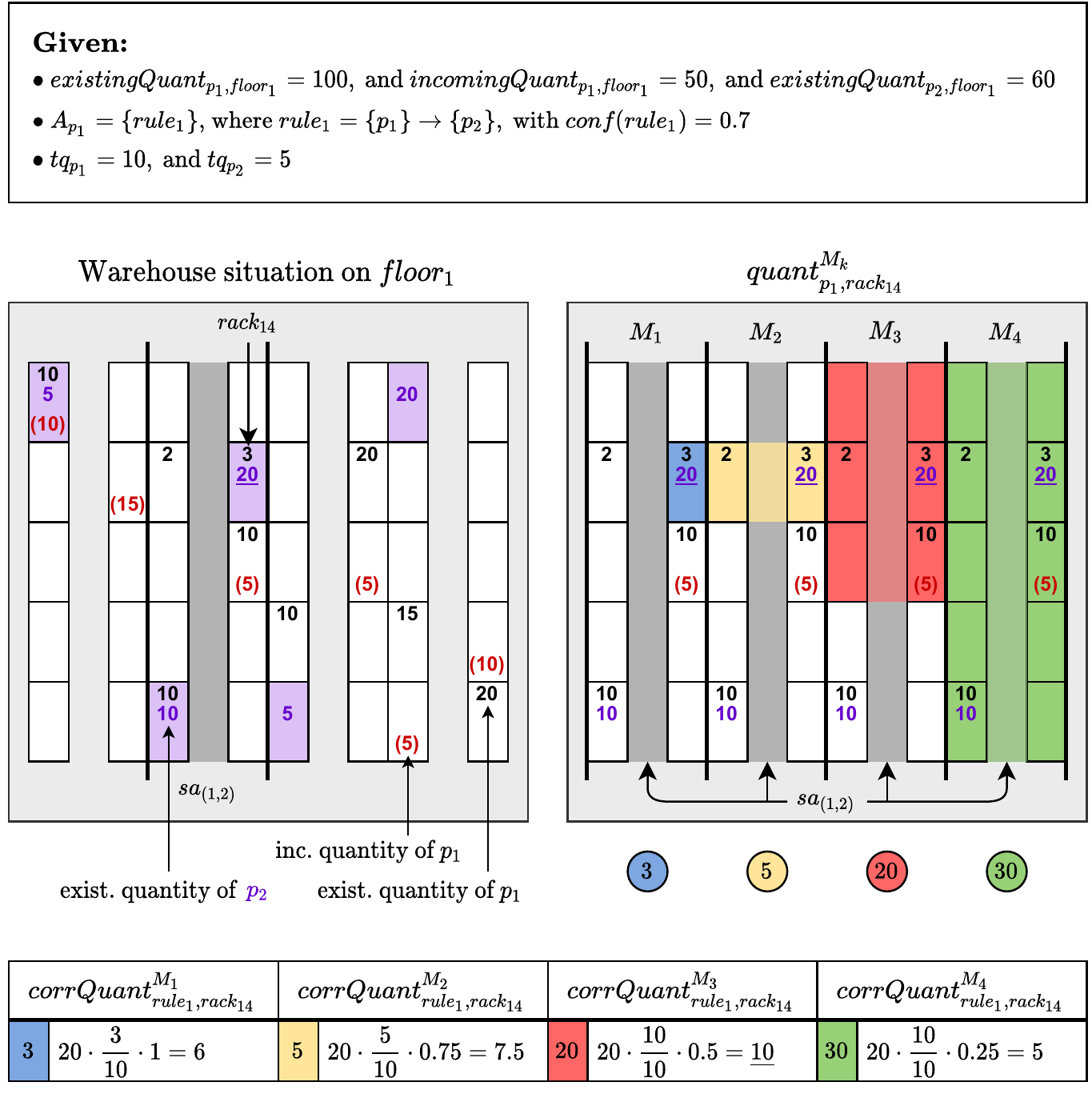}
\caption{Calculating the quantity of the correlated product $p_2$ that is available in the vicinity of the incoming product $p_1$.}
\label{fig:correlationscore}
\end{figure*}

\subsection{Genetic Operators}
The NSGA-II is a genetic algorithm and requires the definition of selection, crossover, and mutation operators. 

\subsubsection{Selection}
We apply a binary tournament selection operator where two random parent individuals compete against each other~\cite{Deb2002}.
The individual with the higher Pareto rank is declared the winner and is allowed to participate in the crossover procedure.
In case both parents are of equal Pareto rank, the individual with the larger crowding distance, i.e. the higher diversity, wins the tournament.

\subsubsection{Crossover}
Since all chromosomes created during a run of the NSGA-II algorithm are of equal length, we use the traditional single-point crossover operator.
It selects a random crossover point on both parents' chromosomes, splits them, and recombines them cross-wise to obtain two new children.

\subsubsection{Mutation}
We define eight mutation operators that incorporate domain-specific knowledge to guide the search process: 
(1) The \textbf{FillRack} mutator selects a random rack and fills it with incoming items from the same sub aisle.
(2) The \textbf{MoveRack} mutator selects a random rack containing at least one incoming item and moves them to a different rack within the same sub aisle.
(3) The \textbf{FillSubAisle} mutator selects a random sub aisle and fills it with incoming items from other sub aisles until it provides the product's target quantity.
(4) The \textbf{ClearSubAisle} mutator selects a random sub aisle and moves any incoming items to a different sub aisle.
(5) The \textbf{RedistributeExceedingQuantities} mutator redistributes incoming items of racks that provide more items than the target quantity to racks that require only a few items to provide the target quantity.
(6) The \textbf{ShiftRacks} mutator shifts all incoming items towards a randomly selected direction: left, right, up, or down.
(7) The \textbf{SwapSubAisles} mutator first groups the sub aisles into pairs and swaps incoming items randomly within each pair.
(8) The \textbf{SwapRacks} mutator is similar to (7) but swaps items within pairs of racks instead of sub aisles.

\subsection{NSGA-II Algorithm}
The overall procedure of our NSGA-II algorithm is summarized in Algorithm~\ref{alg:nsga2}. 
\begin{algorithm}[htb]
\caption{Proposed NSGA-II Algorithm.}
\label{alg:nsga2}
\SetKwInOut{Parameter}{Parameter}
\DontPrintSemicolon
\LinesNumbered
\KwIn{product, quantity, fittingRacks}
\Parameter{parentPopSize, mutProb, $L$, $\delta_{lim}$, maxGen}
\KwOut{paretoFront}
\BlankLine
	pop$_{parent}$ = initParentPopulation(product, quantity, fittingRacks, parentPopSize)\;
	gen = 0\;
	historyOfMaxCD = new List()\;
	\While{gen $<$ maxGen \&\& $std(L) > \delta_{lim}$ }{
    	gen++\;
		pop$_{children}$ = createChildrenPopulation(pop$_{parent}$, ...)\;
		pop$_{combined}$ = pop$_{parent}$ $\cup$ pop$_{children}$\;
		pop$_{parent}$  = createNextParentPopulation(pop$_{combined}$, parentPopSize)\;
		paretoFront = calculateParetoFront(pop$_{parent}$)\;
		maxCD = calculateMaxCD(paretoFront)\;
		historyOfMaxCD.add(maxCD)\;
    }
        \KwRet calculateParetoFront(pop$_{parent}$)\;
\end{algorithm}
The algorithm receives the \texttt{product} to be stored and its \texttt{quantity} as well as the list \texttt{fittingRacks}.
Further, the \texttt{parentPopSize} defines the size of the parent population, the mutation probability is given by \texttt{mutProb}, the number of generations to be used when calculating the standard deviation of the maximum crowding distance~$std(L)$ is called $L$, the threshold for the standard deviation of the crowding distance is $\delta_{lim}$, and the maximum number of generations is called \texttt{maxGen}.
In the end, the algorithm returns a \texttt{paretoFront} of the best storage assignments.

In the first step, the algorithm initializes the population by randomly creating the required amount of chromosomes. 
Therefore, the algorithm selects fitting racks for the product randomly which might produce invalid solutions due to exceeded rack spaces.
Each invalid chromosome is then repaired by moving the amount of exceeding products to another available rack.
Then, the generation counter \texttt{gen} and the history of observed maximum crowding distances are initialized. 
Then, the while loop starts and iterates using the two following stopping criterions:
(i)~the number of maximum generations~(\texttt{maxGen}) is executed, or
(ii)~the standard deviation of observed crowding distances~($std(L)$) falls below the given threshold~($\delta_{lim}$). 
Inside the while loop, the generations counter is incremented, and a complete new children population in the size of the parent population is bred using the proposed selection, crossover and mutation operators.~(\texttt{createChildrenPopulation}).
This set is added to a combined population of existing parent individuals and select the best individuals to fill the new parent population~(\texttt{createNextParentPopulation()}).
Afterwards, a Pareto front is calculated from this parent population~(\texttt{calculateParetoFront()}) and the maximum crowding distance of this front is calculated.
This value is added to the history of maximum crowding distances.
In case, the while loop stops, the current Pareto front is returned.

Since the NSGA-II algorithm returns a Pareto front, a user is usually required to identify the most valuable trade-off solution.
However, we automate this step by applying the following procedure.
For each of the four objective functions~($of_i$), we select the solution~($s_j$) of the Pareto front with the highest value~($of_i(s_j)$) for this function.
We then use these values as a 4-dimensional reference point~($p_{ref} = [e_1, e_2,e_3,e_4]$).
Based on the Euclidean distance, the solution that is closest to the reference point is automatically selected as the most valuable trade-off solution.

\section{Order Picking}
\label{sec:orderpicking}
This section introduces our order picking approach that is based on Ant Colony Optimization.
The overall goal of this algorithm is to construct a pick route for a given customer order.
Since the travel distance is an essential economic goal, the pick route should be as short as possible.
Additionally, the pick route should also be ergonomically favorable. 
The need for changing floors should be minimal to reduce the order picker's physical stress.
Further, the product picking sequence is relevant as if light products are picked first, the order picker might need to rearrange the already picked products so that light products are placed on top of heavy products.
Hence, the order picking algorithm aims to construct a short pick route that collects heavy products first and changes floors as little as possible to address economic and ergonomic criteria.

The main idea of this approach is to represent a mezzanine warehouse as a graph and let ants search for satisfactory order picking sequences.
We make the following assumptions to better deal with the complexity of the order picking problem:
(i)~The state of the mezzanine warehouse does not change while the algorithm is running, that is no repositioning or removal of products is performed.
(ii)~The start and ending p/d points of a pick route may differ.
(iii)~Narrow sub aisles may only be traversed to the sub aisles' midpoint, as the picker always must go back to the cart in the wide pick aisle.
(iv)~The order pickers visit only one rack each time they enter a sub aisle.
(v)~Picking carts withstand infinite loads and can carry an unlimited amount of items.

\subsection{Constraints}
For the order picking algorithm, we define a set of hard and soft constraints.
The hard constraints assess the feasibility of a solution, while the soft constraints measure the extent to which the solution fulfills economic and ergonomic goals.
We define the following hard constraints:
The pick route must start and end at a p/d~point~($HC_1$).
The pick route must collect the requested quantities of the products specified in the pick list~($HC_2$).
After entering a narrow sub aisle, the route must always return to the sub aisle's entrance~($HC_3$).
Further, we define one economic soft constraint: 
The travel distance should be minimal~($SC_1$); 
And two ergonomic soft constraints:
The need for changing floors should be minimal~($SC_2$).
Heavy products should be picked first, followed by lighter products~($SC_3$).

\subsection{Graph Representation}
\begin{figure*}[htb]
\centering
\includegraphics[width=0.7\textwidth]{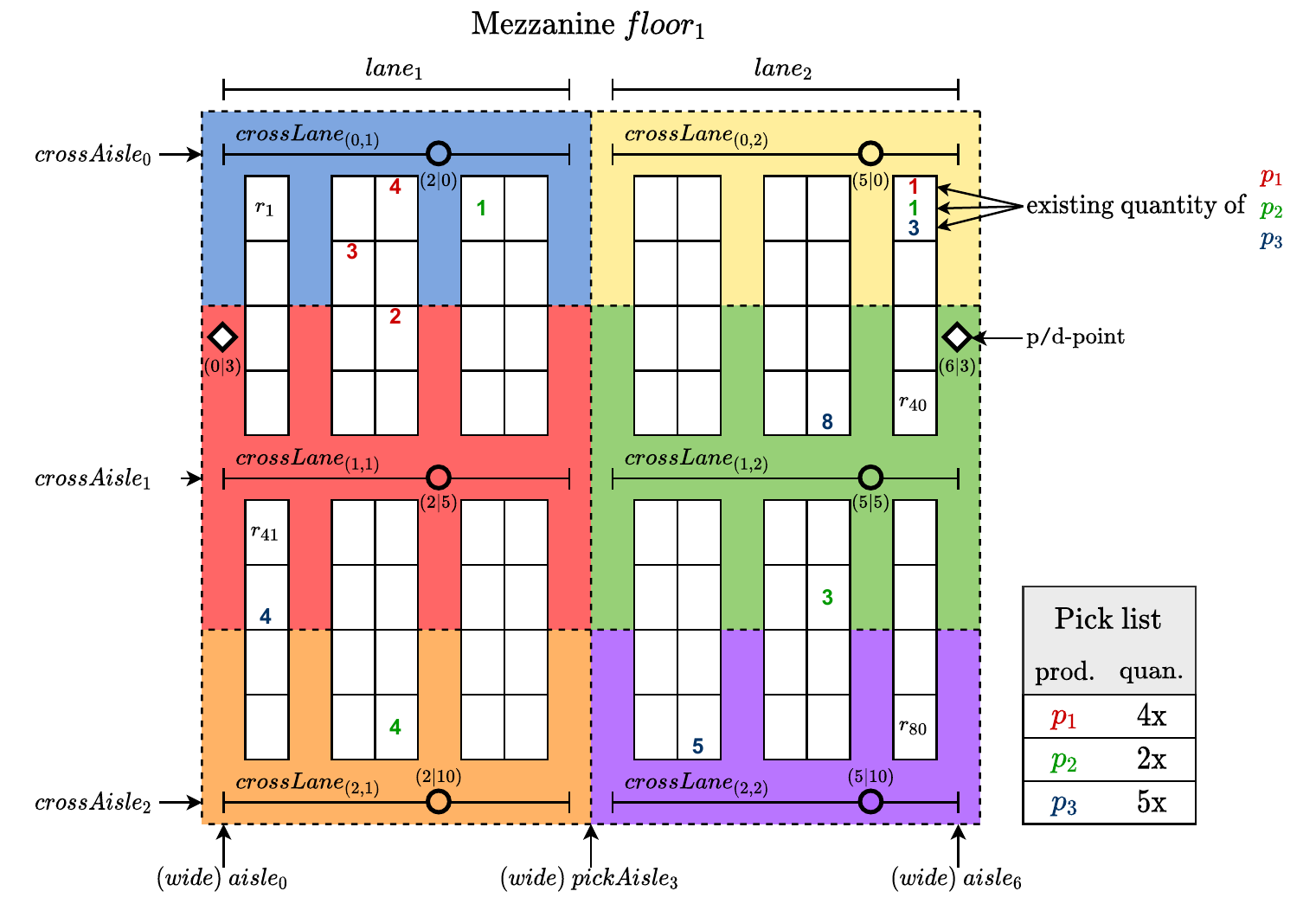} 
\caption{The floor is divided into multiple market zones.}
\label{fig:graphRepresentation1}
\end{figure*}

We propose the following procedure for transferring a mezzanine warehouse into a graph representation.
Figure~\ref{fig:graphRepresentation1} illustrates the procedure of dividing the warehouse into multiple zones called market zones.

Each market zone is represented by a market, and thus, a node in the graph.
The figure depicts the state of $floor_1$ that consists of three cross aisles, three wide (pick) aisles, and two p\texttt{/}d-points. 
We define six market zones obtained by dividing the floor along the wide (pick) aisles into multiple vertical lanes.
In the depicted example, $lane_1$ refers to the area from $aisle_0$ to $pickAisle_3$, and $lane_2$ refers to the area from $pickAisle_3$ to $aisle_6$.
A $crossLane_{(c,l)}$ refers to the part of the cross aisle $c$ that lies within the lane~$l$.
For each cross lane, we define a market zone that comprises the storage racks that can be visited from the respective cross lane up to their midpoints.
For example, the red market zone includes the racks that can be visited if the order picker is located at the $crossLane_{(1,1)}$.
The market zones are limited to the midpoints of the corresponding sub aisles, which prevents the ants from constructing pick routes that entirely traverse the pick aisles.
A market is referred to as $market_{(f, c, l)}$, where $f$ denotes the floor, $c$ the cross aisle, and $l$ the lane.
For each market, we define three attributes: 
(1) the market's \textit{coordinates}, (2) the market's \textit{closest p}\texttt{/}\textit{d-point}, and (3) the market's \textit{supply} that specifies which products are available at which quantity.  
After defining all markets, they are connected via edges to create a complete directed graph.
The edges' weights represent the Manhattan distances between the markets.
If the warehouse consists of a second $floor_2$, the markets on $floor_1$ are also connected to the markets on $floor_2$ and vice versa, with an extra $floorPenalty$ added to the edges' weights.

\subsection{Pick Route Construction}
An ant colony explores the graph to construct a set of pick routes, i.e., a sequence of markets that provide the products, for a given pick list.
A pick route consists of two layers: (i) representing markets, and (ii) rack sequences.

\begin{figure*}[htb]
\centering
\includegraphics[width=0.7\textwidth]{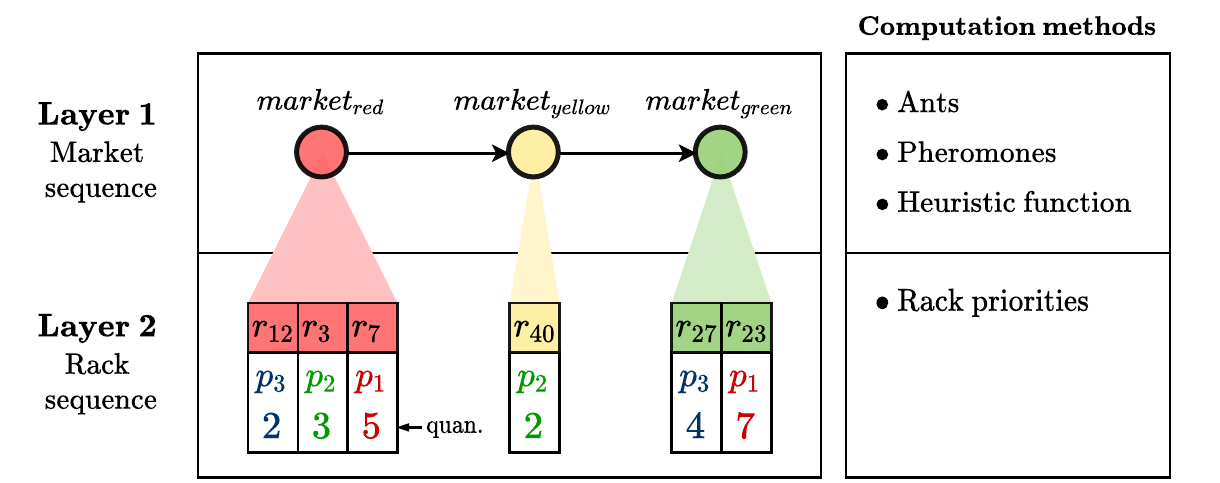} 
\caption{A pick route consists of a market sequence and a rack sequence.}
\label{fig:routeConstruction1}
\end{figure*}

Figure~\ref{fig:routeConstruction1} depicts an example pick route created by a single ant of the colony.
The market sequence~(layer one) of a pick route is computed by an ant that is placed on a market within the graph.
Guided by the pheromone trails, the ant visits neighboring markets until it collected the requested product quantities specified in the pick list.
The ant manages a purchasing list that specifies the missing items.
The pick route is complete after the ant's purchasing list is empty.
Further, each ant must decide whether it enters the market zone from the left or from the right side which depends on the position of the previously visited market. 
The left/right entrance is located at the position where the cross lane has its lowest/highest x coordinate value.
The decision from which side the ant enters the market zone depends on the position of the previously visited market.
While constructing pick routes, the ant applies a heuristic function to identify the markets within its vicinity that seem attractive to visit next.

The second layer represents the rack sequence, i.e., the racks the ant visited in each market.
When calculating the rack sequence, the ants use the following priority rules:
(1)~Racks that provide heavy products should be visited first.
(2)~Racks located closer to the sub aisle's entrance should be visited second.
(3)~Racks that provide the largest quantities should be visited third.

\subsection{Heuristic Function}
To identify the most promising paths and assess the attractiveness of a market, the ants apply a heuristic function.
The attractiveness of a market is based on two factors: (i) the closeness of the market to the ant's current location, and (ii) the availability of required items.
Thus, we define the heuristic function as follows:
\begin{equation} 
\label{eq:heuristicFunction}
\eta_{m,n}^k = \left( \frac{1}{d_{m,n}} \right) \left(I_n^k \right) \text{ , where }n \in U^k
\end{equation}
where $\eta_{m,n}^k$ is the heuristic value that the ant $k$ currently located at market~$m$ associates with the edge $(m,n)$ leading to market $n$.
$U^k$ is the set of markets the ant has not visited yet and $d_{m,n} > 0$ refers to the Manhattan distance between the markets.
$I_n^k \in [0;1]$ denotes the percentage to which the required items of ant $k$ are available at market $n$. 
The higher the heuristic value, the more attractive is the market for the ant.

\subsection{Objective Functions}
After retrieving possible pick routes from the algorithm, we use two objective functions to asses the quality of the route.

\subsubsection{Travel Distance}
This objective function calculates the travel distance of a pick route and measures the extent to which the soft constraint $SC_1$ and $SC_2$ are satisfied.
We define a pick route $P$ to be $P = (M, R)$ where $M = (m_1, ..., m_k)$ refers to the market sequence and $R = (r_1, ..., r_l)$ refers to the rack sequence.
We then define the objective function as follows:
\begin{align*} 
\label{eq:travelDistance2}
travelDistance(P) &= d_{m_1}^{pd} + \sum_{i=1}^l d_{r_i}^{sub} + \sum_{i=1}^k d_{m_i}^{cross} \\
&+ \sum_{i=1}^{k-1} d_{(m_i, m_{i+1})}^{market} + d_{m_k}^{pd}
\end{align*}
where we sum up the distance from the start p\texttt{/}d-point to the first market, the sum of the distances within each entered sub aisle~($d_{r_i}^{sub}$), the sum of the distances within the cross lanes~($d_{m_i}^{cross}$), the distances between the visited markets~($d_{(m_i, m_{i+1})}^{market}$), and the distance from the last visited market to its closest p\texttt{/}d-point~($d_{m_k}^{pd}$).

\subsubsection{Weight Violation}
The second objective function measures the extent to which a pick route satisfies the soft constraint $SC_3$ and counts the number of weight violations in the product picking sequence.
A weight violation occurs if a heavy product is collected after a much lighter product.
In this case, the order picker must rearrange the lighter products already placed on the picking cart to prevent damage.
The user-specified threshold \texttt{allowedWeightDifference} defines the acceptable weight difference between the heavier and the lighter products.
Using this threshold, we count the number of weight violations in a product picking sequence.

\subsection{ACO Algorithm Procedure}
This section proposes our proposed ACO algorithm and shows the pseudo-code in Algorithm~\ref{alg:opMain}.
First of all, the algorithm constructs the graph and initializes the pheromones.
The pheromones are initialized with their maximum possible value determined by $\tau_{max}$.
Additionally, a minimum pheromone can be specified by using the value $\tau_{min}$ in the parametrization of the algorithm.
Then, a while loop starts and uses the concept of cataclysms~\cite{Chen2013} and a maximum number of iterations as stopping criterion:
The parameter \texttt{maxCataclysms} specifies the maximum number of cataclysms that may occur.
The parameter \texttt{maxconsIterWoImpr} defines the time window in which the ACO algorithm must improve the current Pareto front to prevent the cataclysm operator from being applied. 
The parameter \texttt{maxIter} defines the maximum allowed number of iterations regardless of happened cataclysms.

Inside the loop the number of current iterations is incremented and pick routes are constructed.
The general idea is to place one ant on each market of the graph from which the ant starts to create a pick route.
The next market is selected based on the pheromone values and the heuristic function.
We propose two different versions of the ACO to combine these values as explained later.
For each found pick route, the reverse pick route is calculated by reversing the market sequence, toggling the sides from which the ant entered the markets, and recalculating the rack sequence.
We store the pick routes the ants construct in each iteration in the variable \texttt{pickRoutes}.
In the next step, the Pareto-optimal pick routes of this iteration are selected by calculating the objective function and the Pareto rank of all routes.
Afterward, the iteration-best~(\texttt{pickRoutes}$_{\texttt{ib}}$) and the global-best pick routes~(\texttt{nextPickRoutes}$_{\texttt{gb}}$) are merged into a single set and the Pareto-optimal pick routes in this set represent the next set of global-best pick routes.
The iteration-best pick routes and the global-best pick routes are used to perform the pheromone update, which is explained later.
In the further course of the iteration, the ACO algorithm checks whether the cataclysm operator must be applied and compares the global best pick routes of the last and the current iteration.
If the ACO algorithm succeeded in improving the Pareto front, the set \texttt{pickRoutes}$_{\texttt{gb}}$ is updated, and the counter variable \texttt{consIterWoImpr} is reset to 0.
However, if no improvement was made, this counter variable is incremented.
If multiple consecutive iterations fail to achieve an improvement, the search is considered stuck, and the cataclysm operator is applied.
In case the cataclysm is applied, the global-best pick routes \texttt{pickRoutes}$_{\texttt{gb}}$ are included in the set \texttt{pickRoutes}$_{\texttt{cataclysm}}$, the pheromones on the edges representing the pick routes in \texttt{pickRoutes}$_{\texttt{gb}}$ are reset to the lowest possible value, and the set \texttt{pickRoutes}$_{\texttt{gb}}$ is emptied.
Then, the number of cataclysms is incremented and the counter variable \texttt{consIterWoImpr} is reset to 0.
After the main loop terminates, the global-best pick routes \texttt{pickRoutes}$_{\texttt{gb}}$ of the last iteration are included in the set \texttt{pickRoutes}$_{\texttt{cataclysm}}$ and the algorithm returns the Pareto-optimal pick routes in this set.

\begin{algorithm}[h!]
\caption{Proposed ACO Algorithm.}
\label{alg:opMain}
\SetKwInOut{Parameter}{Parameter}
\DontPrintSemicolon
\LinesNumbered
\KwIn{warehouseState, pickList}
\Parameter{maxIterWoImpr, maxCataclysms, maxIter}
\KwOut{pickRoutes}

\BlankLine
	graph = constructGraph()\\
	pheromones = initializePheromones()\\
	\While{cataclysms $<$ maxCataclysms $\mid\mid$ iter $<$ maxIter}{
		iter++\\
		pickRoutes = constructPickRoutes()\\
		pickRoutes$_{ib}$ = selectParetoPickRoutes(pickRoutes)\\
		pickRoutes$_{merged}$ = pickRoutes$_{ib}$ $\cup$ pickRoutes$_{gb}$\\
		nextPickRoutes$_{gb}$ = selectParetoPickRoutes(pickRoutes$_{merged}$)\\
		updatePheromones()\\
    	\uIf{isParetoFrontImproved()}{
			pickRoutes$_{gb}$ = nextPickRoutes$_{gb}$\\
			consIterWoImpr = 0
		}
		\Else{
			consIterWoImpr++\\
			\If{consIterWoImpr $>=$ maxIterWoImpr}{
				pickRoutes$_{cataclysm}$ = pickRoutes$_{cataclysm}$ $\cup$ pickRoutes$_{gb}$\\
				resetPheromonesOnGlobalBestRoutes()\\
				cataclysms++\\
				consIterWoImpr = 0\\
			}
		}
	}
	pickRoutes$_{cataclysm}$ = pickRoutes$_{cataclysm}$ $\cup$ pickRoutes$_{gb}$\\
    	\KwRet selectParetoOptimalPickRoutes(pickRoutes$_{cataclysm}$)\;
\end{algorithm}

\subsection{ACO$_3$ Variant}
In the following, we introduce two variants of our algorithm that show a distinct pheromone handling.
Both variants are inspired by~\cite{Alaya2007} that propose four different variants to handle multi-objective problems with an ACO. 
We select the two best performing variants~(ACO$_3$ and ACO$_4$) and integrate them in our approach to compare which variant produces the best results in our problem domain.
This section introduces the ACO$_3$ variant that applies one ant colony using a single pheromone matrix~$\tau^1$ for optimizing both objectives simultaneously.
In each construction step, the probability of selecting an edge calculates as:
\begin{equation} 
\label{eq:opPheromoneFactor1}
prob_{m,n}^k = \frac{(\tau_{m,n}^1)^{\alpha}(\eta_{m,n}^k)^{\beta}}{\sum\limits_{u \in U^k}(\tau_{m,n}^1)^{\alpha}(\eta_{m,n}^k)^{\beta}}\text{, where }n \in U^k
\end{equation}
where $prob_{m,n}^k$ denotes the probability of ant $k$ located at market $m$ to select the edge $(m,n)$ leading to market $n$.
$\tau_{m,n}^1$ refers to the pheromone value of edge $(m,n)$.
$\eta_{m,n}^k$ denotes the heuristic value (see Formula~\ref{eq:heuristicFunction}) that the ant associates with the edge $(m,n)$.
The parameters $\alpha$ and $\beta$ control the importance of the pheromone values and heuristic values.
Lastly, $U^k$ represents the set of markets that ant~$k$ has not visited yet.

When performing the pheromone update, the ACO$_3$ variant rewards in 90\% of the time the iteration-best pick routes and in 10\% of the time, the global-best pick routes (found since the last cataclysm) to update the pheromone matrix $\tau^1$.
The pheromone values are updated according to the following rule~\cite{Alaya2007}:
\begin{equation}
    \label{eq:opPheromoneUpdate1}
\tau_{m,n}^1 = (1-\rho) \cdot \tau_{m,n}^1 + \Delta \tau_{m,n}^1
\end{equation}
\begin{equation}
\Delta \tau_{m,n}^1 = 
\begin{cases}
    1,		\text{ if } (m,n) \text{ belongs to a pick route in }PF \\
    0,               	\text{ otherwise}
\end{cases}
\end{equation}
where $\rho$ refers to the evaporation factor and $\Delta \tau_{m,n}^1$ is the amount of pheromone that is added to the edge $(m,n)$.
$PF$ refers to the Pareto front containing the solutions to be rewarded.

\subsection{ACO$_4$ Variant}
The ACO$_4$ variant also applies one ant colony but a pheromone matrix~$\tau^1$ for optimizing the first objective function, and another pheromone matrix $\tau^2$ for optimizing the second objective function.
When deciding which edge to explore next, an ant randomly chooses a pheromone matrix.
In each construction step, the probability of selecting an edge calculates as~\cite{Alaya2007}:
\begin{equation} 
\label{eq:opPheromoneFactor2}
p_{m,n}^k = \frac{(\tau_{m,n}^i)^{\alpha}(\eta_{m,n}^k)^{\beta}}{\sum\limits_{u \in U^k}(\tau_{m,n}^i)^{\alpha}(\eta_{m,n}^k)^{\beta}} \text{, where } n \in U^k \text{ and } i \in \{1, 2\}
\end{equation}
where $\tau_{m,n}^r$ refers to the pheromone value of edge $(m,n)$ w.r.t. pheromone matrix $\tau^i$.
At the end of an iteration, the ACO$_4$ variant updates the pheromone matrix $\tau^i$ by rewarding the iteration-best pick route $PR_{ib}^i$ that minimizes the objective function $of_i$~\cite{Alaya2007}:
\begin{equation} 
\label{eq:opPheromoneUpdate2}
\tau_{m,n}^i = (1-\rho) \cdot \tau_{m,n}^i + \Delta \tau_{m,n}^i 
\end{equation}
\begin{equation}
\Delta \tau_{m,n}^i = 
\begin{cases}
    \frac{1}{1 + of_i(PR_{ib}^i)-of_i(PR_{gb}^i)},&\parbox[t]{2cm}{if $(m,n)$ belongs to the pick route $PR_{ib}^i$} \\
    0,               &\text{otherwise}
\end{cases}
\end{equation}
where $\rho$ again refers to the evaporation factor and $\Delta \tau_{m,n}^i$ is the pheromone added to the edge $(m,n)$ in pheromone matrix~$\tau^i$.
$PR_{gb}^i$ refers to the global-best pick route that minimizes the $i$th objective function of all pick routes constructed since the last cataclysm occurred.

\section{Evaluation}
\label{sec:eval}
This section presents the evaluation of our approaches.
It defines the used warehouse models for applying our algorithms, presents performance indicators, summarizes alternative policies to which we compare our algorithms, and provides the parameter settings of our algorithms.
Afterwards, we first evaluate our storage assignment and order picking algorithms individually before we evaluate the interaction of both algorithms.

\subsection{Mezzanine Warehouse Models}
\begin{figure*}[hbt]
\centering
\includegraphics[width=0.7\textwidth]{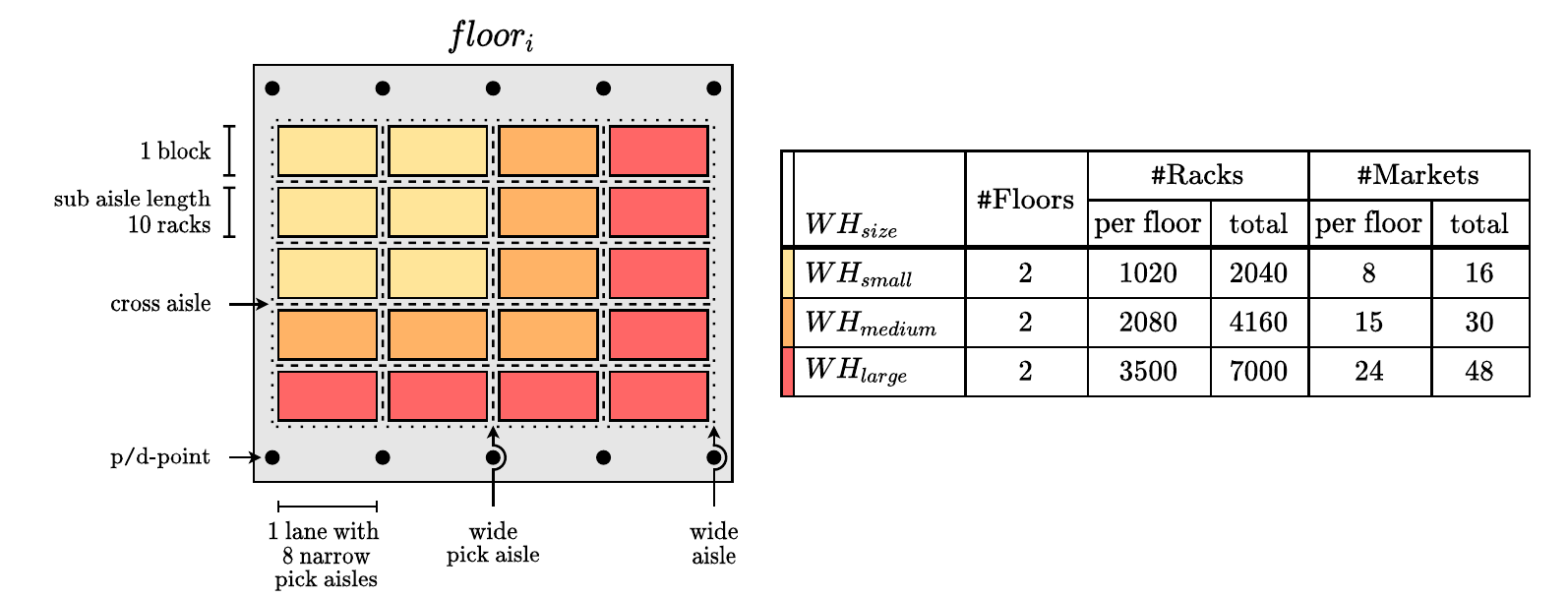}
\caption{The warehouses $WH_{small}, WH_{medium}$, and $WH_{large}$ use different floor layouts.}
\label{fig:warehouseSizes}
\end{figure*}

The NSGA-II and the ACO algorithm are evaluated in three artificial mezzanine warehouses of different sizes that are defined in cooperation with our cooperation company to build real-world test cases.
The warehouses are shown in Figure~\ref{fig:warehouseSizes}: $WH_{small}$~(yellow), $WH_{medium}$~(orange), and $WH_{large}$~(red).
For the small, medium, and large warehouses, we define the size of the product assortment to be 500, 1000, and 1500, respectively.
Since each product requires a weight, we define three normal distributions and a probability to determine the weight using this distribution: 25\% to use $\mathcal{N}(2, 1.0^2)$, 50\% to use $\mathcal{N}(5, 2.0^2)$, and 25\% to use $\mathcal{N}(8, 1.0^2)$.
Using these distributions and probabilities, we aim at a representative set of product weights where most of the products have a medium weight and some products have low and some have heavy weights.
The products might also have correlations to up to three other products: With a probability of 30\%, 40\%, 20\%, and 10\% a product has no, one, two, or three correlated products, respectively, with a randomly generated correlation confidence between 10\% and 90\%. 
For evaluating the order picking algorithm, we fill the storage up to 50\% of the available storage space and randomly generate 100 customer orders based on the product assortment and given correlations between products.
Each customer order comprises 20 items to pick that are selected as follows: 
We split the product assortment into four equally sized groups based on the product rank. 
With a probability of 40\%, 30\%, 20\%, and 10\% an order contains an item of the highest, second highest, third highest, and lowest rank class, respectively, which ensures that high-ranked products appear more often in customer orders.

\subsection{Performance Indicators for Assessing Pareto Fronts}
Since we assess a multi-objective optimization problem, the algorithms compute a Pareto front.
We use the following quality indicators for Pareto fronts introduced by~\cite{Wang2016}.
The Coverage~(C) quality indicator quantifies the extent to which a computed Pareto front covers the reference Pareto front.
The quality indicators Generational Distance~(GD) and Euclidean Distance~(ED) measure the distance from a computed Pareto front to a reference Pareto front or a reference solution.
The quality indicators Pareto Front Size~(PFS) and Generated Spread~(GS) measure the diversity of the solutions that exist in a computed Pareto front.
Finally, the quality indicator Inverted Generational Distance~(IGD) combines convergence and diversity aspects.
Since most of these quality indicators require the calculation of a reference Pareto front, we use the Pareto fronts returned by all algorithms as basis.
From these Pareto fronts, we select the non-dominated solutions of the union of all computed Pareto fronts and use this front as reference Pareto front.

\subsection{Alternative Policies}
We use the following alternative policies for the storage assignment problem.
The \textbf{random storage assignment policy} allocates the incoming items to random racks on the floors in clusters of target quantity size~\cite{Bartholdi2019}.
In the \textbf{closest open location storage assignment policy}, the warehouse employees select the storage locations for storing an incoming product, which are usually the racks closest to the p\texttt{/}d-points~\cite{Dekoster2007}.
The \textbf{rank-based storage assignment policy} assigns fast-moving products close to the p\texttt{/}d-points, while slow-moving products are assigned to racks further away~\cite{Petersen1999}.

For the order picking problem, we apply a modified S-Shape heuristic for comparison that constructs \mbox{s-shaped} pick routes based on the graph representation~\cite{Petersen1997}.
This heuristic uses all markets as starting point iteratively as well as the reversed versions of each route to generate a Pareto front of possible solutions.

\subsection{Algorithm Parameter Settings}
Based on a preliminary parameter study, we parameterize our \mbox{NSGA-II} algorithm as follows:
We set the mutation probability to 0.95 for all warehouse sizes so that the mutation operators are applied very frequently. 
Further, we define parameters dependent on the warehouse size (small/medium/large): 
The parent population size is set to~(50/60/70), and the maximum number of generations to~(200/250/300).
These values increase with the size of the warehouse since the number of possible solutions increases with the warehouse size and we provide the algorithm more exploration possibilities~(population size) and more time~(number of generations) for optimizing the solutions.

Further, we set the parameters for our ACO algorithm as follows:
In line with the literature, we set the pheromone factor~$\alpha$ to 1.0 and the heuristic factor~$\beta$ to 2.0.
We set the evaporation factor~$\rho$ to 0.02 causing the pheromones to evaporate rather slowly which enables the algorithm to achieve a higher degree of exploration especially in the early stages.
The min/max values for the pheromone matrices~($\tau_{min/max}$) are set to 1 and 25, respectively, add a floor change penalty of 50, and set the allowed weight difference to 3\,kg.
As stopping criterion, we set the maximum number of cataclysms to 3 and hence, the algorithm terminates after it became stuck for the third time.
Using the results of a preliminary parameter study, we set the maximum consecutive iterations without improvements to 20, and the maximum iterations to 250 since these values yield the best results w.r.t the scores.

\subsection{Evaluation of the NSGA-II Algorithm for Storage Assignment Tasks}
We evaluate our NSGA-II algorithm against the random, closest open location, and rank-based storage assignment policies.
We apply all approaches on the three warehouse sizes~(Setting 1.a, 1.b, 1.c) and on five randomly generated storage assignment tasks, i.e., we select a random product from the product assortment and set the quantity to be assigned to the quantity already existing in the warehouse.
We repeat the execution of the NSGA-II algorithm ten times to reduce random effects and present mean and standard deviation values.
All generated solutions of all algorithms are then used to calculate the reference Pareto front required for the quality indicators.
Table~\ref{tab:samean} summarizes the mean values and Table~\ref{tab:sasd} shows the standard deviation values for this evaluation.
The coverage~(C) results for the small warehouse show that the NSGA-II Pareto front covers about~90\% of the reference Pareto front while the other approaches cover only around 9\% and 1\%
Since, the NSGA-II shows the lowest GD and ED values, this Pareto front is located closest to the reference front.
The NSGA-II algorithm finds around 48 solutions per problem instance with a maximum possible value of 50 solutions for the small warehouse size.
The other policies only construct 22 to 28 Pareto-optimal solutions while their maximum possible value is set to 500.
Further, the NSGA-II achieves the lowest GS and IGD values which indicates, that the solutions converge well towards the reference Pareto front and offer diverse solutions.
\begin{table}[tb]
	\centering
	\caption{
	Mean values of the six quality indicators achieved by the storage assignment algorithms in Setting 1.a, 1.b, and 1.c (best values are shown in bold).}
	\label{tab:samean}
	\begin{adjustbox}{width=\columnwidth}
	\begin{tabular}{c p{1.3cm} *{6}{c}} \toprule
	Setting & Policy & C [$\mu$]  & GD [$\mu$]  & ED [$\mu$]  & PFS [$\mu$]& GS [$\mu$]  & IGD [$\mu$]  \\	\midrule
	\multirow{4}{*}{\parbox[c]{1cm}{\centering 1.a}} & Random & 0.01 &  1.59 & 25.33 &  24.80  & 0.73& 2.26\\
    & Closest & 0.01 & 2.14 & 28.37  & 21.98  & 0.74  & 2.53  \\
	& Rank & 0.09 & 0.95  & 21.88  & 28.20  & 0.78  & 2.53 \\
	& NSGA-2 & \textbf{0.90} & \textbf{0.04}  & \textbf{16.48} & \textbf{47.52} & \textbf{0.50}  & \textbf{0.26}  \\ \midrule
	
	\multirow{4}{*}{\parbox[c]{1cm}{\centering 1.b}} & Random & 0.01  & 2.38 & 26.20  & 16.80 & 0.75  & 1.92  \\
		& Closest & 0.00  & 3.32  & 31.28 & 14.82  & \textbf{0.73}  & 2.18  \\
		& Rank & 0.06  & 1.68 & 22.03  & 19.44  & 0.83  & 2.13 \\
		& NSGA-2 & \textbf{0.93}  & \textbf{0.02}  & \textbf{14.01}  &\textbf{52.84}  & 1.04  & \textbf{0.11} \\ \midrule
		
	\multirow{4}{*}{\parbox[c]{1cm}{\centering 1.c}} & Random & 0.00  & 2.78  & 29.47  & 9.92  & \textbf{0.81}  & 1.86  \\
		& Closest & 0.00 & 3.11  & 28.58  & 11.08  & 0.86  & 2.10  \\
		& Rank & 0.01 & 1.40  & 25.23  & 14.24  & 0.94  & 1.92 \\
		& NSGA-2 & \textbf{0.99} & \textbf{0.00}  & \textbf{16.76}  & \textbf{64.34}  & 1.37 &\textbf{0.02}  \\ \bottomrule
	\end{tabular}
	\end{adjustbox}
\end{table}

\begin{table}[tb]
	\centering
	\caption{
	Standard deviations of the six quality indicators achieved by the storage assignment algorithms in Setting 1.a, 1.b, and 1.c.}
	\label{tab:sasd}
	\begin{adjustbox}{width=\columnwidth}
	\begin{tabular}{c p{1.3cm} *{6}{c}} \toprule
	Setting & Policy &  C [$\sigma$] & GD [$\sigma$] & ED [$\sigma$]& PFS [$\sigma$] & GS [$\sigma$] & IGD [$\sigma$] \\	\midrule
	\multirow{4}{*}{\parbox[c]{1cm}{\centering 1.a}} & Random & 0.01 & 0.78 &  10.20 &  11.52 &  0.16 &  1.91 \\
    & Closest &  0.01  & 1.26  & 12.29 & 9.49  & 0.19  & 1.71 \\
	& Rank &  0.09  & 0.44  & 10.66  & 14.98  & 0.21  & 1.99 \\
	& NSGA-2 & 0.10 &  0.08 & 7.93 &  5.35 &  0.17 & 0.26 \\ \midrule
	
	\multirow{4}{*}{\parbox[c]{1cm}{\centering 1.b}} & Random &  0.02 & 1.21 &  15.20  & 13.29  & 0.13 & 0.96 \\
		& Closest & 0.01 & 1.74  & 18.95 & 9.84 &  0.08 &  0.95 \\
		& Rank & 0.12 &  1.14  & 13.53  & 19.53 & 0.18  & 1.13 \\
		& NSGA-2 &  0.14 & 0.05 & 9.38 & 9.46  & 0.54  & 0.20 \\ \midrule
		
	\multirow{4}{*}{\parbox[c]{1cm}{\centering 1.c}} & Random & 0.00 &  2.34 & 26.32  & 4.89 & 0.13 & 1.81 \\
		& Closest & 0.00 &  3.55  & 26.45 & 7.49 &  0.16 &  2.17 \\
		& Rank &  0.01  & 1.39  & 23.91 & 6.79 & 0.17 & 2.12 \\
		& NSGA-2 & 0.01 & 0.00 &  14.06 &  9.60 & 0.45 & 0.06 \\ \bottomrule
	\end{tabular}
	\end{adjustbox}
\end{table}

\begin{figure*}[htb]
    \centering
    \includegraphics[width=0.8\textwidth]{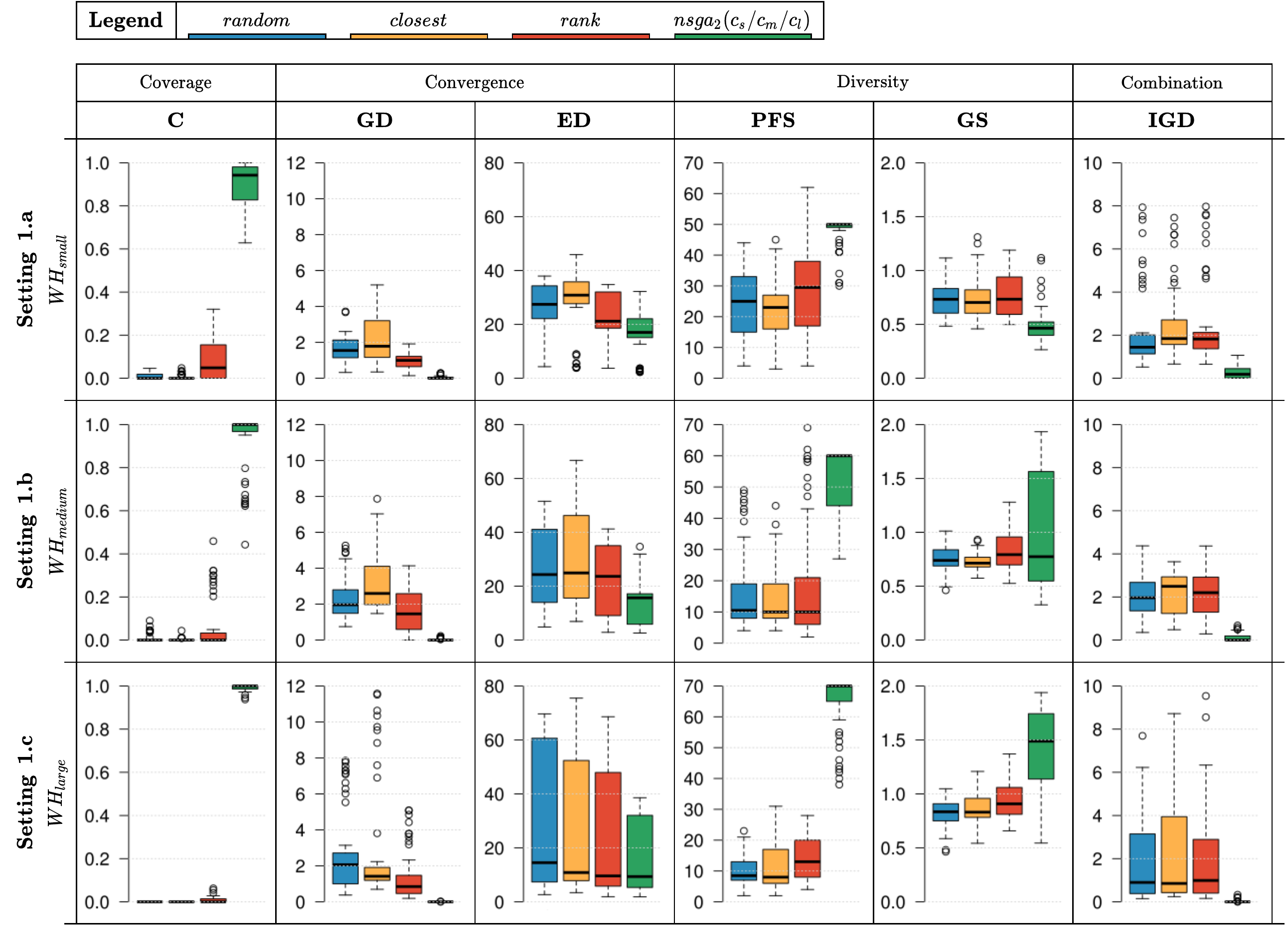}
    \caption{Box plots of the six performance indicators achieved by the storage assignment strategies in Setting 1.a, 1.b, and 1.c.}
    \label{fig:eval:wh_sa_box}
\end{figure*}

In the medium warehouse, the results show similar behavior. 
The table shows that the Pareto front $PF_{nsga_2(c_m)}$ covers approximately 93\% of the reference Pareto front $PF_{ref}$.
Except for some outliers, the alternative policies struggle to cover the solutions in $PF_{ref}$.
The observed GD and ED values are fairly similar to the values in Setting 1.a.
However, the standard deviations of the ED metric increased noticeably, which may be related to the larger search space where the solutions tend to be more spread out.
Nevertheless, the Pareto front $PF_{nsga_2(c_m)}$ still achieves the lowest GD and ED values, indicating that this Pareto front converges best towards $PF_{ref}$.
Concerning the PFS metric, the \mbox{NSGA-II} algorithm finds about 53 solutions per problem instance, while the alternative policies find approximately less than~20~solutions per problem instance.
The GS values of $PF_{nsga_2(c_m)}$ increased remarkably, which may be due to the larger parent population size and the larger search space that make it difficult for the NSGA-II algorithm to fill the gaps in the Pareto front so that all solutions are evenly distributed.
Lastly, the IGD values of $PF_{nsga_2(c_m)}$ are close to 0, indicating that $PF_{nsga_2(c_m)}$ represents the entire reference Pareto front $PF_{ref}$ in most cases.
 
Similarly, the large warehouse shows comparable results.
The Pareto front $PF_{nsga_2(c_l)}$ covers about 99\% of the reference Pareto front $PF_{ref}$, while $PF_{rank}$ covers only 1\%.
Thus, almost all solutions found by the rank-based policy are dominated by the solutions found by the NSGA-II algorithm. 
The mean GD value of $PF_{nsga_2(c_l)}$ equals 0, indicating that the entire Pareto front $PF_{nsga_2(c_l)}$ is part of $PF_{ref}$ in almost all cases.
Other than that, the observations made in the previous Setting 1.b also occur in this setting. 
Thus, the standard deviations of the ED metric further increase, the \mbox{NSGA-II} algorithm finds the most solutions per problem instance, the alternative policies find fewer solutions per problem instances, the GS values of $PF_{nsga_2(c_l)}$ further increase, and the IGD values $PF_{nsga_2(c_m)}$ are closer to 0.

In summary, the results show that the random and the closest open location policy struggle to cover even a single solution in the reference Pareto front. 
Additionally, the NSGA-II algorithm outperforms the alternative policies in smaller warehouses.
Further, the NSGA-II finds on average the most solutions per problem instance and the solutions are less equally distributed.

In addition to the quality evaluation, we also measure the mean execution time of the approaches for solving 50 problem instances in each warehouse size\footnote{We run our experiments on a MacBook Pro using macOS Sierra 10.12.6, a 2.2GHz Intel Core i7 CPU and 16GB DDR3 RAM.}.
The alternative policies achieve low execution times of about 0.17/0.30/0.50 seconds for small/medium/large which is due to their comparably simple operation.
In the warehouse small/medium/large, the NSGA-II algorithm achieves execution times of about 2/6/15 seconds, which is due to the increase population and iteration count for larger warehouses.
The execution times of the NSGA-II algorithm may be considered acceptable, as the algorithm requires only a few seconds to find storage allocations that are remarkably better than the ones found by the alternative policies.

\subsection{Evaluation of the ACO Algorithm for Order Picking Tasks}
We evaluate both versions of our ACO algorithm against the modified \mbox{S-Shape} heuristic.
We apply all approaches on the three warehouse sizes~(Settings 2.a, 2.b, 2.c) using five customer orders randomly selected from the set of generated customer orders as explained earlier and repeat the execution of the ACO algorithms ten times to reduce random effects and present mean and standard deviation values.
Then, we use all generated solutions the algorithms to calculate the reference Pareto front required for the quality indicators.
Table~\ref{tab:opmean} summarizes the mean values and Table~\ref{tab:opsd} shows the standard deviation values for this evaluation. 
For the small warehouse, the Pareto fronts of the $ACO_3$ and $ACO_4$ variants cover 74\% of the reference Pareto front while the S-Shape heuristic fails to cover even a single solution. 
Both ACO algorithms achieve nearly the same GD and ED values and the close to zero GD values show that many solutions are part of the reference front. 
The ACO algorithms find around ten solutions per problem instance, while the S-Shape only finds three solutions per problem instance.
The S-Shape achieves the lowest, hence, the best GS values, but it is not meaningful to compare these values to the ACO ones as it contains only three solutions that are considerably worse than solutions of the ACO algorithms. 
The ACO algorithms achieve low IGD values, indicating that both Pareto fronts converge well towards the reference front and provide diverse solutions.
\begin{table}[tb]
    \centering
	\caption{
	Mean values of the six quality indicators achieved by the order picking algorithms in Setting 2.a, 2.b, and 2.c (best values are shown in bold).}
	\label{tab:opmean}
	\begin{adjustbox}{width=\columnwidth}
	\begin{tabular}{c p{1.4cm} *{6}{c}} \toprule
		Setting & Policy & C [$\mu$]& GD [$\mu$] &  ED [$\mu$]  & PFS [$\mu$]  & GS [$\mu$]  & IGD [$\mu$] \\	\midrule
		
		\multirow{3}{*}{\parbox[c]{1cm}{\centering 2.a}} & sShape & 0.00 & 22.15 & 80.34 & 2.80  & \textbf{0.84}  & 18.28 \\
		& $ACO_3$ & 0.73 & \textbf{1.40}  & 32.03 & \textbf{10.66} &  0.99  & 2.74  \\
		& $ACO_4$ & \textbf{0.74} &  1.59 & \textbf{32.07}  & 9.54 & 0.87 & \textbf{1.91} \\	\midrule

		\multirow{3}{*}{\parbox[c]{1cm}{\centering 2.b}} & sShape & 0.00 & 31.20 &  117.42  & 3.60  & 0.72  & 18.78  \\
		& $ACO_3$ & \textbf{0.69} &  \textbf{1.97} & \textbf{50.11} & \textbf{12.14}  & 0.81 & \textbf{3.00}  \\
		& $ACO_4$ & 0.33 & 4.30 &  54.59 &  11.30 &  \textbf{0.66}  & 4.15\\	\midrule

		\multirow{3}{*}{\parbox[c]{1cm}{\centering 2.c}} & sShape & 0.00 &  41.41 &  121.35  & 2.00  & 0.88  & 27.18  \\
		& $ACO_3$ & \textbf{0.84} & \textbf{1.56}  & \textbf{56.75}  & \textbf{10.50}  & 0.74  & \textbf{5.64}  \\
		& $ACO_4$ & 0.16 & 9.78 & 70.02 & 10.14  & \textbf{0.68} & 7.28  \\		\bottomrule
	\end{tabular}
	\end{adjustbox}
\end{table}

\begin{table}[tb]
    \centering
	\caption{
	Standard deviations of the six quality indicators achieved by the order picking algorithms in Setting 2.a, 2.b, and 2.c.}
	\label{tab:opsd}
	\begin{adjustbox}{width=\columnwidth}
	\begin{tabular}{c p{1.4cm} *{6}{c}} \toprule
		Setting & Policy  & C [$\sigma$]  & GD [$\sigma$] & ED [$\sigma$]  & PFS [$\sigma$]  & GS [$\sigma$]  & IGD [$\sigma$] \\	\midrule
		
		\multirow{3}{*}{\parbox[c]{1cm}{\centering 2.a}} & sShape &  0.00  & 14.20 & 28.26 & 1.60  & 0.16& 6.55 \\
		& $ACO_3$ & 0.13 &  1.91 &  16.21 &  5.63 &  0.36 &  2.96 \\
		& $ACO_4$ &  0.18 &  2.64 & 16.14 & 3.97  & 0.35 & 2.52 \\	\midrule

		\multirow{3}{*}{\parbox[c]{1cm}{\centering 2.b}} & sShape & 0.00 & 6.92 &  34.51  & 1.02  & 0.12  & 5.03 \\
		& $ACO_3$ & 0.15 & 2.08 &  15.74 &  3.80 & 0.20  & 2.67 \\
		& $ACO_4$ & 0.16 & 3.62 & 14.43 & 3.23 & 0.17 & 2.68 \\	\midrule

		\multirow{3}{*}{\parbox[c]{1cm}{\centering 2.c}} & sShape &  0.00 &  16.42 & 35.71 & 0.89 & 0.14  & 6.95 \\
		& $ACO_3$ &  0.11 &  2.90 &  18.95  & 2.87& 0.23  & 6.27 \\
		& $ACO_4$  & 0.11  & 5.83  & 21.76  & 3.28  & 0.21 & 3.91 \\		\bottomrule
	\end{tabular}
	\end{adjustbox}
\end{table}

\begin{figure*}[htb]
    \centering
    \includegraphics[width=0.8\textwidth]{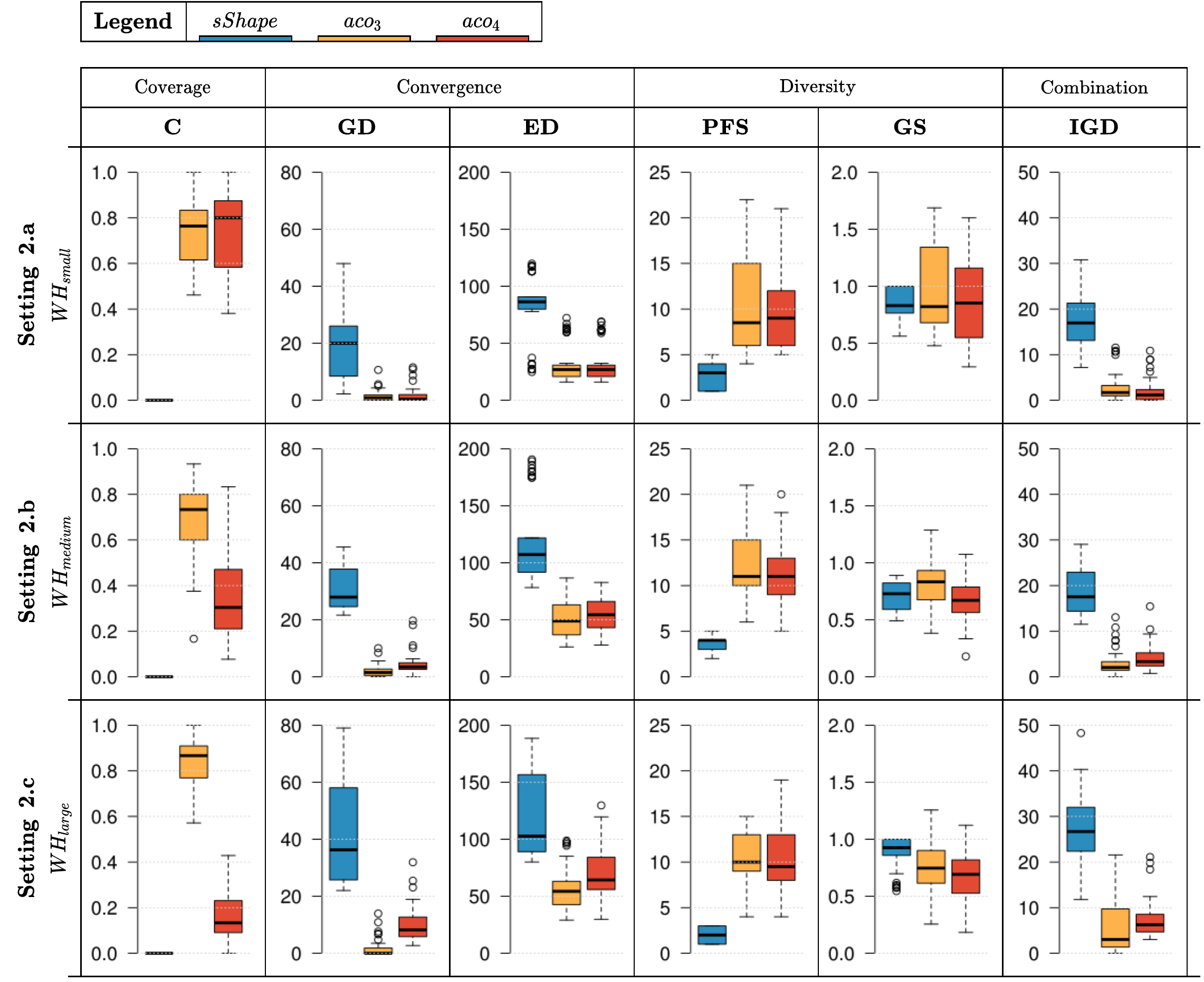}
    \caption{Box plots of the six performance indicators achieved by the order picking strategies in Setting 2.a, 2.b, and 2.c.}
    \label{fig:eval:wh_op_box}
\end{figure*}

The metrics of the medium warehouse show similar behavior as in the previous setting.
Like in the previous setting, the Pareto front $PF_{sShape}$ fails to cover even a single solution in the reference Pareto front $PF_{ref}$.
The Pareto front $PF_{aco_3}$ covers approximately 69\% of $PF_{ref}$, while $PF_{aco_4}$ covers only 33\%.
Thus, the ACO$_3$ variant tends to find better pick routes than the ACO$_4$ variant.
The Pareto front $PF_{aco_3}$ achieves the lowest GD and ED values among all computed Pareto fronts.
Thus, $PF_{aco_3}$ converges best towards $PF_{ref}$, which is not surprising, as $PF_{aco_3}$ covers most of the solutions in $PF_{ref}$.
The GD and ED values of $PF_{aco_4}$ are slightly larger than the ones of $PF_{aco_3}$, indicating that $PF_{aco_4}$ does not converge as well as $PF_{aco_3}$ towards $PF_{ref}$.
Compared to the previous setting, the GD and ED values of $PF_{sShape}$ increased, which may be due to the larger search space.
Concerning the PFS metric, the ACO$_3$ variant finds about 12 solutions per problem instance, followed closely by the ACO$_4$ variant that finds around 11 solutions per problem instance, while the S-Shape heuristic only finds about 4 solutions per problem instance.
Regarding the GS~metric, the solutions in $PF_{aco_4}$ are slightly better distributed than the solutions in $PF_{aco_3}$.
With respect to the IGD metric, $PF_{aco_3}$ achieves the lowest IGD values, indicating that $PF_{aco_3}$ converges well towards $PF_{ref}$ and offers a high diversity of solutions.

In the large warehouse, the S-Shape heuristic is again unable to cover a solution in $PF_{ref}$.
The Pareto front $PF_{aco_3}$ covers 84\% of the reference Pareto front $PF_{ref}$, while $PF_{aco_4}$ covers only 16\%.
Thus, most of the solutions found by the ACO$_4$ variant are dominated by the solutions found by the ACO$_3$ variant.
Accordingly, the ACO$_4$ variant has problems to compete with the ACO$_3$ variant in larger warehouses.
Compared to the previous setting, the GD and ED values of $PF_{aco_4}$ further increased, indicating that the distances between the solutions in $PF_{aco_4}$ and the solutions in $PF_{ref}$ became larger.
The Pareto front $PF_{aco_3}$ converges best towards $PF_{ref}$, as it achieves the lowest GD and ED values. 
Regarding the PFS metric, both ACO variants find about 10 solutions per problem instance.
The GS metric indicates that the solutions in $PF_{aco_4}$ are marginally better distributed than the solutions in $PF_{aco_3}$.
Finally, the Pareto front $PF_{aco_3}$ achieves the lowest, and thus, best IGD values, signalizing that $PF_{aco_3}$ converges best towards $PF_{ref}$ and offers diverse solutions.

In summary, the ACO algorithms outperform the S-Shape heuristic in all warehouse sizes, and $ACO_3$ and $ACO_4$ show similar performance in smaller warehouses.
With increasing warehouse size, the solutions found by the $ACO_3$ variant dominate more and more solutions of the $ACO_4$ variant. 
Hence, the $ACO_3$ variant starts to find better pick routes than the $ACO_4$ variant, while the $ACO_4$ variant produces slightly better distributed solutions.

In addition to the quality evaluation, we also measure the mean execution time of the approaches for solving 50 problem instances in each warehouse size.
The S-Shape heuristics takes around 0.15 seconds to compute routes.
The $ACO_3$ and the $ACO_4$ variant achieve fairly the same execution times in all warehouse sizes of around 1/3/6 seconds for $WH_{small}$/$WH_{medium}$/$WH_{large}$.
As the warehouse size increases, the graph consists of more markets causing more ants to be deployed in each iteration.
Still, we consider the ACO execution times acceptable, as they require only a few seconds to find noticeably better pick routes.

\subsection{Evaluation of the Interaction between NSGA-II and ACO Algorithm}
In the previous section, we have shown the applicability of our algorithms for storage assignment and order picking in dedicated analyses. 
The results indicate that both algorithms outperform state-of-the-art solutions for those tasks.
In this section, we evaluate the interaction between our proposed algorithms by assessing them in three settings:
Section~\ref{sec:nsgaRandomAco3} determines whether the $ACO_3$ performs better on the NSGA-II planned warehouse compared to the random warehouse;
Section~\ref{sec:nsgaRandomAco4} performs a similar assessment for the $ACO_4$;
Section~\ref{sec:nsgaAco} evaluates whether the $ACO_3$ or the $ACO_4$ perform better on the NSGA-II planned warehouse.
Tables~\ref{tab:interactionmean} and \ref{tab:interactionsd} summarize the results.
\begin{table}[tb]
	\centering
	\caption{
	Mean values of the six quality indicators achieved by the combination of storage assignment and order picking algorithm in Setting 3 to 5 (best values are shown in bold).}
	\label{tab:interactionmean}
	\begin{adjustbox}{width=\columnwidth}
	\begin{tabular}{c p{2.8cm} *{6}{c}} \toprule
		Setting & Policy & C [$\mu$] &  GD [$\mu$]  & ED [$\mu$] & PFS [$\mu$] & GS [$\mu$] & IGD [$\mu$]\\	\midrule
	    
	    \multirow{2}{*}{\parbox[c]{1cm}{\centering 3.a}}& Random, $ACO_3$ & 0.00 &  60.32  & 215.43 & 9.96 & \textbf{0.97}  & 56.49  \\
		& NSGA-2, $ACO_3$ & \textbf{1.00} & \textbf{0.00} & \textbf{22.62}  & \textbf{12.20} & 1.12  & \textbf{0.00} \\ \midrule

		\multirow{2}{*}{\parbox[c]{1cm}{\centering 3.b}} & Random, $ACO_3$ & 0.00 & 38.37  & 168.05 & \textbf{13.76} & \textbf{0.91} & 38.85 \\
		& NSGA-2, $ACO_3$ & \textbf{1.00} & \textbf{0.00} & \textbf{36.65}  & 12.12 & 0.96  & \textbf{0.00} \\	\midrule

		\multirow{2}{*}{\parbox[c]{1cm}{\centering 3.c}} & Random, $ACO_3$ & 0.00  & 50.06  & 213.69  & \textbf{13.96}  & 0.90  & 51.35  \\
		& NSGA-2, $ACO_3$ & \textbf{1.00} & \textbf{0.00} & \textbf{32.78}  & 12.02 & \textbf{0.87} & \textbf{0.00}  \\	\midrule \midrule 
		
		\multirow{2}{*}{\parbox[c]{1cm}{\centering 4.a}} & Random, $ACO_4$ & 0.00 &  54.03  & 165.75 & 8.14 &  \textbf{0.93} & 46.16  \\
		& NSGA-2, $ACO_4$ & \textbf{1.00}  & \textbf{0.00}  & \textbf{25.08}& \textbf{10.26} & 0.96 & \textbf{0.00} \\ \midrule 
		
		\multirow{2}{*}{\parbox[c]{1cm}{\centering 4.b}} & Random, $ACO_4$ & 0.01 &  44.61 & 198.83 &  \textbf{11.82}  & 0.86 & 48.94  \\
		& NSGA-2, $ACO_4$ & \textbf{0.99}  & \textbf{0.12} & \textbf{35.65}  & 11.10  & \textbf{0.71}  & \textbf{0.21}  \\ \midrule 
		
		\multirow{2}{*}{\parbox[c]{1cm}{\centering 4.c}} & Random, $ACO_4$ & 0.00 &  51.19 &  207.95  & \textbf{12.04}  & 0.85 & 52.91 \\
		& NSGA-2, $ACO_4$ & \textbf{1.00} &  \textbf{0.00} & \textbf{38.08}  & 10.04 &  \textbf{0.71}  & \textbf{0.03}  \\ \midrule \midrule 
		
		\multirow{2}{*}{\parbox[c]{1cm}{\centering 5.a}} & NSGA-2, $ACO_3$ & \textbf{0.80}  & \textbf{1.43}  & \textbf{25.30} & \textbf{10.26}  & 0.99  & 1.38  \\
		& NSGA-2, $ACO_4$ & 0.66 & 1.52 & 26.58 & 9.32 &  \textbf{0.88}  & \textbf{0.98}  \\ \midrule 

        \multirow{2}{*}{\parbox[c]{1cm}{\centering 5.b}} & NSGA-2, $ACO_3$ & \textbf{0.69}  & \textbf{1.40}  & \textbf{21.37}  & \textbf{9.84} & 0.94 & 3.53  \\
		& NSGA-2, $ACO_4$ & 0.41  & 3.83  & 22.85  & 8.14  & \textbf{0.79} & \textbf{3.13}  \\ \midrule 

        \multirow{2}{*}{\parbox[c]{1cm}{\centering 5.c}} & NSGA-2, $ACO_3$ & \textbf{0.79} & \textbf{1.92}  & \textbf{33.80} & 9.16  & 0.81 & \textbf{3.11} \\
		& NSGA-2, $ACO_4$ & 0.22 & 8.69 & 43.25 & \textbf{9.18}  & \textbf{0.66}  & 5.71  \\ \bottomrule

	\end{tabular}
	\end{adjustbox}
\end{table}
\begin{table}[tb]
	\centering
	\caption{
    Standard deviations of the six quality indicators achieved by the combination of storage assignment and order picking algorithm in Setting 3 to 5.}
	\label{tab:interactionsd}
	\begin{adjustbox}{width=\columnwidth}
	\begin{tabular}{c p{2.8cm} *{6}{c}} \toprule
		Setting & Policy  & C [$\sigma$] & GD [$\sigma$] & ED [$\sigma$]  & PFS [$\sigma$]  & GS [$\sigma$]  & IGD [$\sigma$] \\	\midrule
	    
	    \multirow{2}{*}{\parbox[c]{1cm}{\centering 3.a}}& Random, $ACO_3$ &  0.00  & 22.94 & 62.38  & 4.10  & 0.06  & 22.96 \\
		& NSGA-2, $ACO_3$ &  0.00 &  0.00  & 7.51 & 6.17  & 0.35 & 0.00 \\ \midrule

		\multirow{2}{*}{\parbox[c]{1cm}{\centering 3.b}} & Random, $ACO_3$  & 0.00 & 13.66 & 35.09  & 3.88 & 0.07  & 14.41 \\
		& NSGA-2, $ACO_3$ & 0.00 &  0.00  & 16.50  & 4.18  & 0.36  & 0.00 \\	\midrule

		\multirow{2}{*}{\parbox[c]{1cm}{\centering 3.c}} & Random, $ACO_3$ &  0.00 &  14.03  & 54.74  & 3.56  & 0.07 & 17.53 \\
		& NSGA-2, $ACO_3$ &  0.00 & 0.00 &  8.19  & 5.40 & 0.32 &  0.00 \\	\midrule \midrule 
		
		\multirow{2}{*}{\parbox[c]{1cm}{\centering 4.a}} & Random, $ACO_4$ &  0.00 &  18.52  & 73.60 & 3.69  & 0.10  & 26.01 \\
		& NSGA-2, $ACO_4$ & 0.00 &  0.00 &  8.71 & 3.65 & 0.31  & 0.00 \\ \midrule 
		
		\multirow{2}{*}{\parbox[c]{1cm}{\centering 4.b}} & Random, $ACO_4$ &  0.03  & 15.52 & 62.11  & 2.96 & 0.09 & 22.17 \\
		& NSGA-2, $ACO_4$ &  0.03  & 0.57 & 22.53 & 4.20  & 0.28  & 0.77 \\ \midrule 
		
		\multirow{2}{*}{\parbox[c]{1cm}{\centering 4.c}} & Random, $ACO_4$ &  0.01  & 21.77  & 68.40 & 3.55  & 0.08 & 26.20 \\
		& NSGA-2, $ACO_4$ &  0.01 &0.00 &  12.91 & 3.82 & 0.28 & 0.22 \\ \midrule \midrule 
		
		\multirow{2}{*}{\parbox[c]{1cm}{\centering 5.a}} & NSGA-2, $ACO_3$ & 0.15 &  2.70 &  6.61 &  4.74 &  0.20 & 1.93 \\
		& NSGA-2, $ACO_4$ &  0.20  & 2.11 & 7.98  & 3.72  & 0.20 & 0.93 \\ \midrule 

        \multirow{2}{*}{\parbox[c]{1cm}{\centering 5.b}} & NSGA-2, $ACO_3$ &  0.16 & 1.74 &  4.58 &  3.43 & 0.25  & 3.46 \\
		& NSGA-2, $ACO_4$  & 0.16  & 5.59  & 4.58  & 1.90  & 0.22  & 3.48 \\ \midrule 

        \multirow{2}{*}{\parbox[c]{1cm}{\centering 5.c}} & NSGA-2, $ACO_3$ & 0.14 & 3.81 &  12.68 & 3.28  & 0.25 & 4.40 \\
		& NSGA-2, $ACO_4$ &  0.14  & 6.32 & 17.01  & 3.54  & 0.18  & 3.54 \\ \bottomrule
	\end{tabular}
	\end{adjustbox}
\end{table}

\subsubsection{Comparison of NSGA-II and Random Planned Warehouses for $ACO_3$}
\label{sec:nsgaRandomAco3}
In this setting, we apply the $ACO_3$ algorithm on all warehouse sizes~(Setting 3.a, 3.b, 3.c) twice: once for the warehouse that used the NSGA-II algorithm for storage assignment and once for the randomly assigned warehouse.
Again, we select five random items from the product assortment and set the amount to assign to the already existing amount inside the warehouse.
In the small warehouse, the Pareto front of the NSGA-II planned warehouse covers the entire reference front while the random planned warehouse does not cover a single solution in the reference front, hence, the GD and IGD values of the NSGA-II planned warehouse are 0 and the ED values are minimal. 
The high GD and ED values of the random planned warehouse indicate that its Pareto front does not converge well towards the reference front.
Thus, the solutions found in the random warehouse are considerably worse than the solutions found in the NSGA-II warehouse.
In the medium warehouse, the Pareto fronts of random and NSGA-II planned warehouses achieve fairly the same quality indicator values as in the previous setting.
However, the GD and ED metric indicate that the results for the random warehouse unexpectedly converge better towards the reference front than in Setting 3.a.
This could be due to the limited amount of executed problem instances and needs to be further assessed with a higher number of problem instances.
Nevertheless, the Pareto front of the random warehouse is still far from converging towards reference front.
In the large warehouse, the same observations can be made as in the previous settings, underlining that the $ACO_3$ variant finds better pick routes in the NSGA-II warehouse than in the random warehouse.
In summary, the evaluation results show that the NSGA-II algorithm and the $ACO_3$~variant interact well together and the $ACO_3$~variant profits from the NSGA-II algorithm that ensures our four economic constraints.

\begin{figure*}[htb]
    \centering
    \includegraphics[width=0.8\textwidth]{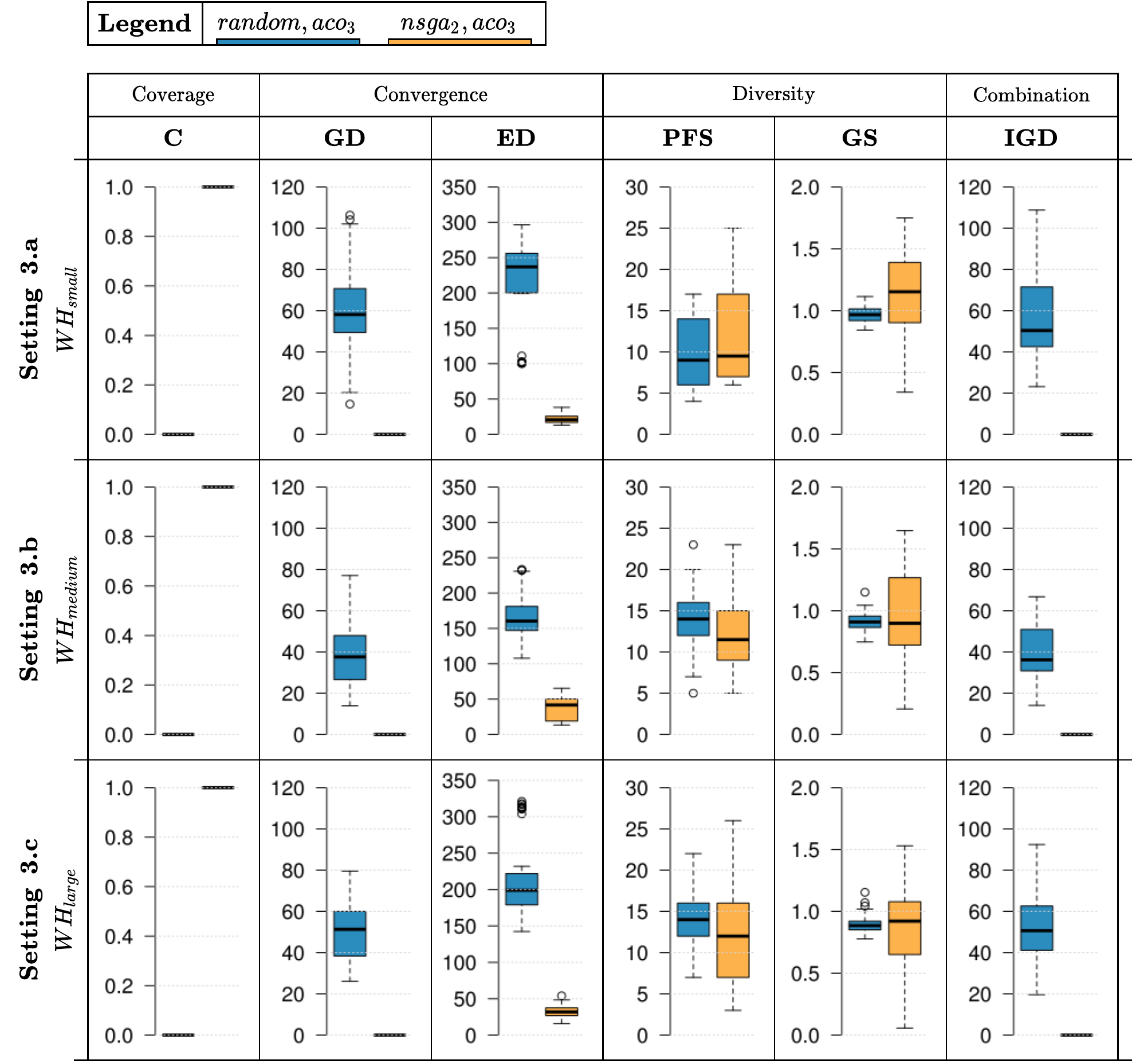}
    \caption{Box plots of the six performance indicators achieved by the interaction evaluation for the ACO$_3$ algorithm in Setting 3.a, 3.b, and 3.c.}
    \label{fig:eval:wh:inter3_box}
\end{figure*}

\subsubsection{Comparison of NSGA-II and Random Planned Warehouses for $ACO_4$}
\label{sec:nsgaRandomAco4}
This setting repeats the Settings 3.a to 3.c for the $ACO_4$ algorithm.
We now discuss the results for the Settings 4.a to 4.c.
In the small warehouse, the Pareto front of the NSGA-II warehouse covers the entire reference Pareto front and the random warehouse fails to cover a single solution.
Thus, all solutions found in the random warehouse are dominated by the solutions of the NSGA-II warehouse.
The GD and ED values for the random warehouse are higher than the ones of the NSGA-II warehouse which shows that the Pareto front of the random warehouse is further away from the reference front. 
Hence, the solutions of the random warehouse are noticeably worse than the solutions found in the NSGA-II warehouse.
In the medium warehouse, the Pareto front of the NSGA-II warehouse does not always cover the entire reference front, while the random warehouse covers at least one solution in the reference front in 7 of 50 repetitions.
Thus, the $ACO_4$ variant occasionally finds a few solutions in the random warehouse that are comparable with the solutions found in the NSGA-II warehouse.
Despite these few outliers, the results show a similar behavior as in the previous setting.
Similar to the previous settings, the evaluation in the large warehouse show comparable results.
The NSGA-II warehouse covers all solutions in the reference front in 49 of 50 repetitions and the random warehouse manages to cover at least one solution in the reference front. 
In summary, the results show that the $ACO_4$ variant also finds better pick routes if the warehouse applies the NSGA-II storage strategy and both algorithms interact well with each other.

\begin{figure*}[htb!]
    \centering
    \includegraphics[width=0.8\textwidth]{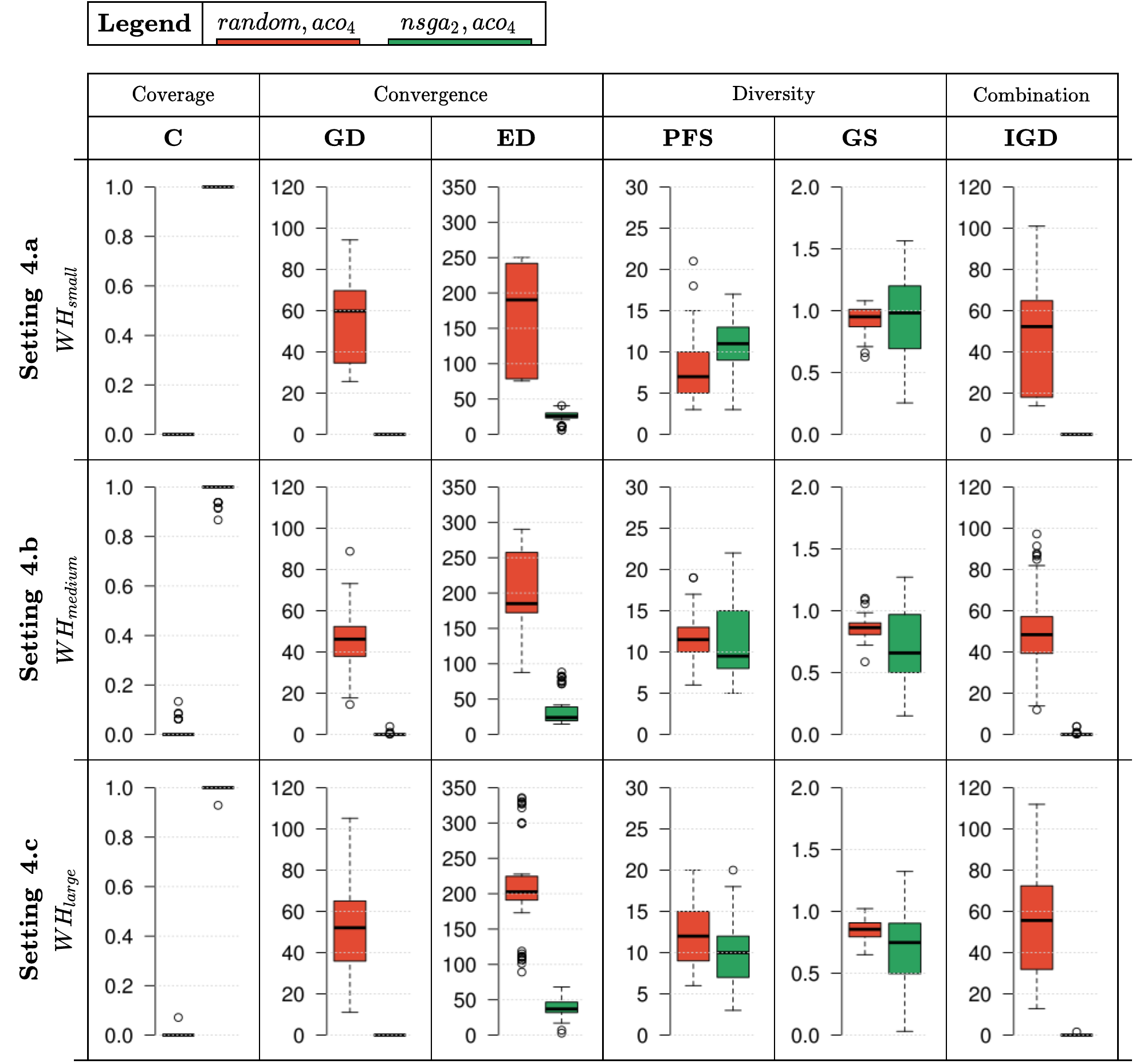}
    \caption{Box plots of the six performance indicators achieved by the interaction evaluation for the $ACO_4$ algorithm in Setting 4.a, 4.b, and 4.c.}
    \label{fig:eval:wh:inter4_box}
\end{figure*}

\subsubsection{Comparison of $ACO_3$ and $ACO_4$ on NSGA-II Planned Warehouses}
\label{sec:nsgaAco}
This section investigates which ACO variant performs better if the warehouse applies the NSGA-II storage strategy.
Both variants are applied on five randomly selected customer orders on all warehouse sizes~(Settings 5.a, 5.b, 5.c). 
In the small warehouse, the Pareto front of the $ACO_3$ algorithm covers approximately 80\% of the reference front, while $ACO_4$ covers only 66\%.
Both ACO variants converge well towards the reference front as indicated by the low GD and ED values and find approximately ten solutions per problem instance, and IGD values of both fronts are almost the same.
However, the GS values indicate, that the solutions of the $ACO_4$ variant have a better distribution than the ones of $ACO_3$.
In the medium warehouse, the Pareto front of $ACO_3$ covers approximately 69\% of the reference front, while the one from $ACO_4$ covers only~41\%, and thus, the solutions found by $ACO_4$ tend to be dominated by the ones from $ACO_3$.
The GD and ED metric indicate that the $ACO_3$ Pareto front converges better towards the reference front.
Similar to the small warehouse, the GS indicate, that solutions of the $ACO_4$ Pareto front are better distributed.
In the large warehouse, the $ACO_3$ dominates $ACO_4$ even more with regards to the Coverage metric.
Furthermore, the GD and ED values of $ACO_4$ increased, indicating that the distance between the solutions in $ACO_4$ Pareto front and the solutions in the reference front become larger.
Again, the solutions in the $ACO_4$ Pareto front have a slightly better distribution than the solutions in the $ACO_3$ Pareto front.
However, this time, $ACO_3$ achieves better IGD values, as $ACO_3$ covers large parts of the reference front.
In summary, we can state that with increasing warehouse size, the $ACO_3$ variant finds better pick routes than the $ACO_4$ variant while both variants find approximately the same number of solutions per problem instance.
However, the solutions found by the $ACO_4$ variant are slightly better distributed than the ones from the $ACO_3$ variant.

\begin{figure*}[htb!]
    \centering
    \includegraphics[width=0.8\textwidth]{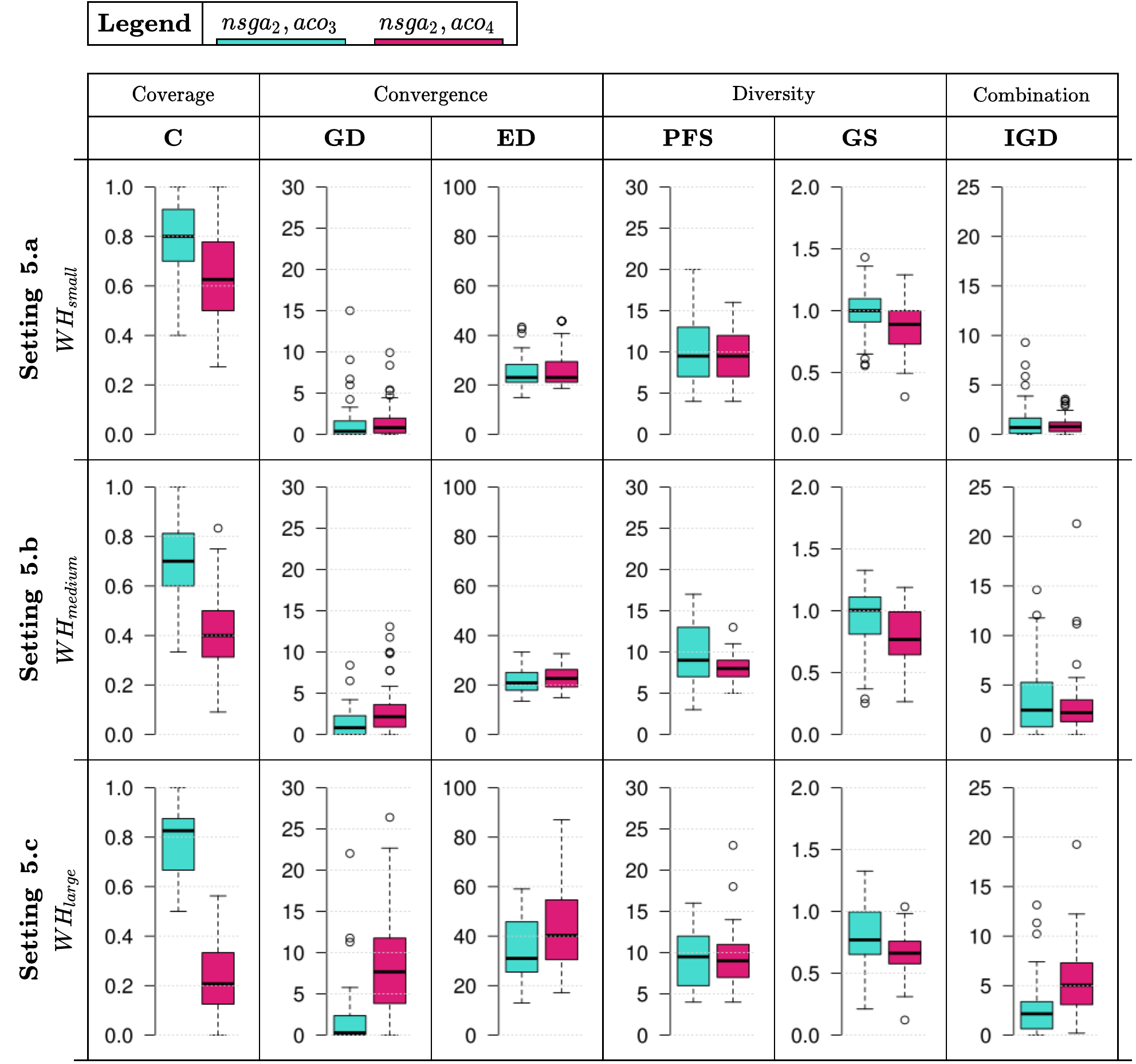}
    \caption{Box plots of the six performance indicators achieved by the interaction evaluation between $ACO_3$ and $ACO_4$ with the NSGA-II algorithm in Setting 5.a, 5.b, and 5.c.}
    \label{fig:eval:wh:inter5_box}
\end{figure*}

\subsection{Threats to Validity}
We identified the following threats to validity of our evaluation.
First, the NSGA-II and ACO algorithms are evaluated in three mezzanine warehouses of different sizes.
However, real-world mezzanine warehouses may consist of more floors, blocks, pick aisles, and racks than specified in the warehouses used for evaluation. 
Nevertheless, we are convinced that our defined warehouses form a representative set for mezzanine warehouses and can easily be extended for further evaluation runs.
Second, since the algorithms are evaluated in warehouses that apply either the random or the NSGA-II storage strategy, the evaluation results may not be transferable to warehouses that apply different storage strategies.
Even though the product assortment, the product correlations, the customer orders, and the storage allocations are randomly generated reflecting specific characteristics of real-world mezzanine warehouses, the proposed algorithms are easily transferable to real application data. 
Third, we decided to compare the ACO algorithm only to one order picking policy. 
This decision was made in awareness of the limited expressiveness of our results but was necessary as the majority of policies in the literature violate assumptions made in this work.
Finally, we only evaluate our NSGA-II and ACO algorithms against heuristic policies.
Hence, they also should be evaluated against other optimization methods like other evolutionary optimization algorithms or graph-based optimization techniques.
However, we decided to do this evaluations as future work.

\section{Conclusion}
\label{sec:conclusion}
Due to the complexity of the storage assignment and the order picking problem, efficient optimization algorithms are required to find satisfactory solutions within reasonable times.
This thesis proposes an NSGA-II algorithm for optimizing the storage assignment problem, and an ACO algorithm for optimizing the order picking problem in mezzanine warehouses.
The algorithms incorporate knowledge about the interdependency between both problems to improve the overall warehouse performance.
Besides optimizing economic constraints, the algorithms also optimize ergonomic criteria, as mezzanine warehouses represent labor-intensive working environments in which the employees account for a large part of the warehouse performance. 
We evaluate the NSGA-II algorithm against three storage assignment policies frequently applied in practice: the random, the closest open location, and the rank-based policy.
The evaluation results show that the NSGA-II algorithm outperforms the alternatives already in smaller warehouses and the larger the warehouse, the better the NSGA-II algorithm prevails against the alternative policies.
We evaluate the ACO algorithm against the S-Shape heuristic that is frequently applied in practice.
Our evaluation results show that the ACO outperforms the S-Shape heuristic in all tested warehouse sizes.
Finally, we evaluate the interaction between the NSGA-II and the ACO algorithm.
The evaluation results show that both ACO variants find better pick routes if the warehouse assigns its products by applying the NSGA-II algorithm instead of the random storage strategy, thus, the NSGA-II and the ACO algorithm interact well with each other.

In the future, we plan to integrate additional features to further increase the applicability of the storage assignment.
First, we want to allow state changes of the mezzanine warehouses while the storage assignment is running which would make some of the solutions in the Pareto front infeasible.
Further, we plan to parallelize the storage assignment algorithm so that the sequential assignment of products is replaced and the execution times will decrease.
Regarding the order picking algorithm, we also aim at parallelizing the execution to reduce the required calculation times. 
Finally, we want to research on integrating forecasts of future assignment and order picking tasks to proactively replace goods within the storage that will be ordered in the near future.

\bibliographystyle{IEEEtran}
\bibliography{Bibliography.bib} 

\end{document}